\documentclass[]{bytedance_seed}

\usepackage[toc,page,header]{appendix}

\usepackage{minitoc}
\usepackage{amssymb}
\usepackage[table]{xcolor}
\usepackage{enumitem}
\usepackage{longtable}
\usepackage{booktabs}
\usepackage{caption}
\usepackage{graphicx}
\usepackage{wrapfig}
\usepackage[utf8]{inputenc}  
\usepackage{float}  
\usepackage{minted}  
\usepackage{listings}
\usepackage{xcolor} 
\usepackage{textcomp}  
\usepackage{footmisc}
\usepackage{tabularx}
\usepackage[most]{tcolorbox} 
\tcbuselibrary{listings}     

\newtcblisting{jsonbox}[1][]{
  listing only,
  listing engine=listings,
  title={#1}, 
  colback=white, 
  colframe=gray!25!black, 
  sharp corners=downhill, 
  boxrule=0.5pt,
  left=5pt, top=5pt, bottom=5pt,
  listing options={
    language=Java, 
    basicstyle=\small\ttfamily,
    keywordstyle=\color{blue!70!black}\bfseries,
    stringstyle=\color{green!50!black},
    commentstyle=\color{gray},
    numberstyle=\tiny\color{gray},
    numbers=left,
    breaklines=true,
    showstringspaces=false,
    morekeywords={true, false, null} 
  }
}

\definecolor{papergreen}{HTML}{008000} 
\definecolor{paperred}{HTML}{D00000}   

\newcommand{\goodres}[2]{%
    #1 {\color{papergreen}\small (+#2)}%
}
\newcommand{\badres}[2]{%
    #1 {\color{paperred}\small (-#2)}%
}
\newcommand{\neutralres}[2]{%
    #1 {\color{gray}\small (#2)}%
}
\makeatletter
\newcommand{\tablestyle}[2]{%
  \setlength{\tabcolsep}{#1}%
  \renewcommand{\arraystretch}{#2}%
  \centering%
  \footnotesize%
}
\makeatother

\definecolor{navyblue}{HTML}{0071BC}
\definecolor{oai-green-200}{HTML}{D7EADF} 
\definecolor{oai-green-400}{HTML}{A6D3B8} 
\definecolor{oai-green-600}{HTML}{63AC86} 
\definecolor{oai-gray-300}{HTML}{D9D9D9}  
\definecolor{oai-gray-600}{HTML}{808080}  
\definecolor{L1_Purple}{HTML}{E6E0F8} 
\definecolor{L2_Orange}{HTML}{FCEFD8} 
\definecolor{L3_Green}{HTML}{EFF8E0}  
\definecolor{L4_Blue}{HTML}{E0F0F8}   

\title{SpatialTree: How Spatial Abilities Branch Out in MLLMs}

\author[\blacktriangle,\bigstar,*]{Yuxi Xiao}
\author[\diamondsuit,\bigstar,*]{Longfei Li}
\author[\bigstar]{Shen Yan}
\author[\bigstar]{Xinhang Liu}
\author[\blacktriangle]{Sida Peng}
\author[\diamondsuit]{\\ Yunchao Wei}
\author[\blacktriangle]{Xiaowei Zhou}
\author[\bigstar, \dagger]{Bingyi Kang}

\affiliation[\blacktriangle]{Zhejiang University}
\affiliation[\bigstar]{ByteDance Seed}
\affiliation[\diamondsuit]{Beijing Jiaotong University}

\contribution[\dagger]{Project Lead}
\contribution[*]{Equal Contribution}

\abstract{
Cognitive science suggests that spatial ability develops progressively—from perception to reasoning and interaction. Yet in multimodal LLMs (MLLMs), this hierarchy remains poorly understood, as most studies focus on a narrow set of tasks. We introduce SpatialTree, a cognitive-science-inspired hierarchy that organizes spatial abilities into four levels: low-level perception (L1), mental mapping (L2), simulation (L3), and agentic competence (L4). Based on this taxonomy, we construct the first capability-centric hierarchical benchmark, thoroughly evaluating mainstream MLLMs across 27 sub-abilities. The evaluation results reveal a clear structure: L1 skills are largely orthogonal, whereas higher-level skills are strongly correlated, indicating increasing interdependency. Through targeted supervised fine-tuning, we uncover a surprising transfer dynamic—negative transfer within L1, but strong cross-level transfer from low- to high-level abilities with notable synergy. Finally, we explore how to improve the entire hierarchy. We find that naïve RL that encourages extensive “thinking” is unreliable: it helps complex reasoning but hurts intuitive perception. We propose a simple auto-think strategy that suppresses unnecessary deliberation, enabling RL to consistently improve performance across all levels. By building SpatialTree, we provide a proof-of-concept framework for understanding and systematically scaling spatial abilities in MLLMs.

}

\date{\today}
\correspondence{Bingyi Kang}
\checkdata[Project Page]{\url{https://spatialtree.github.io/}}

\begin{document}
\maketitle


\begin{figure}[!ht]
\centering
\includegraphics[width=0.98\linewidth]{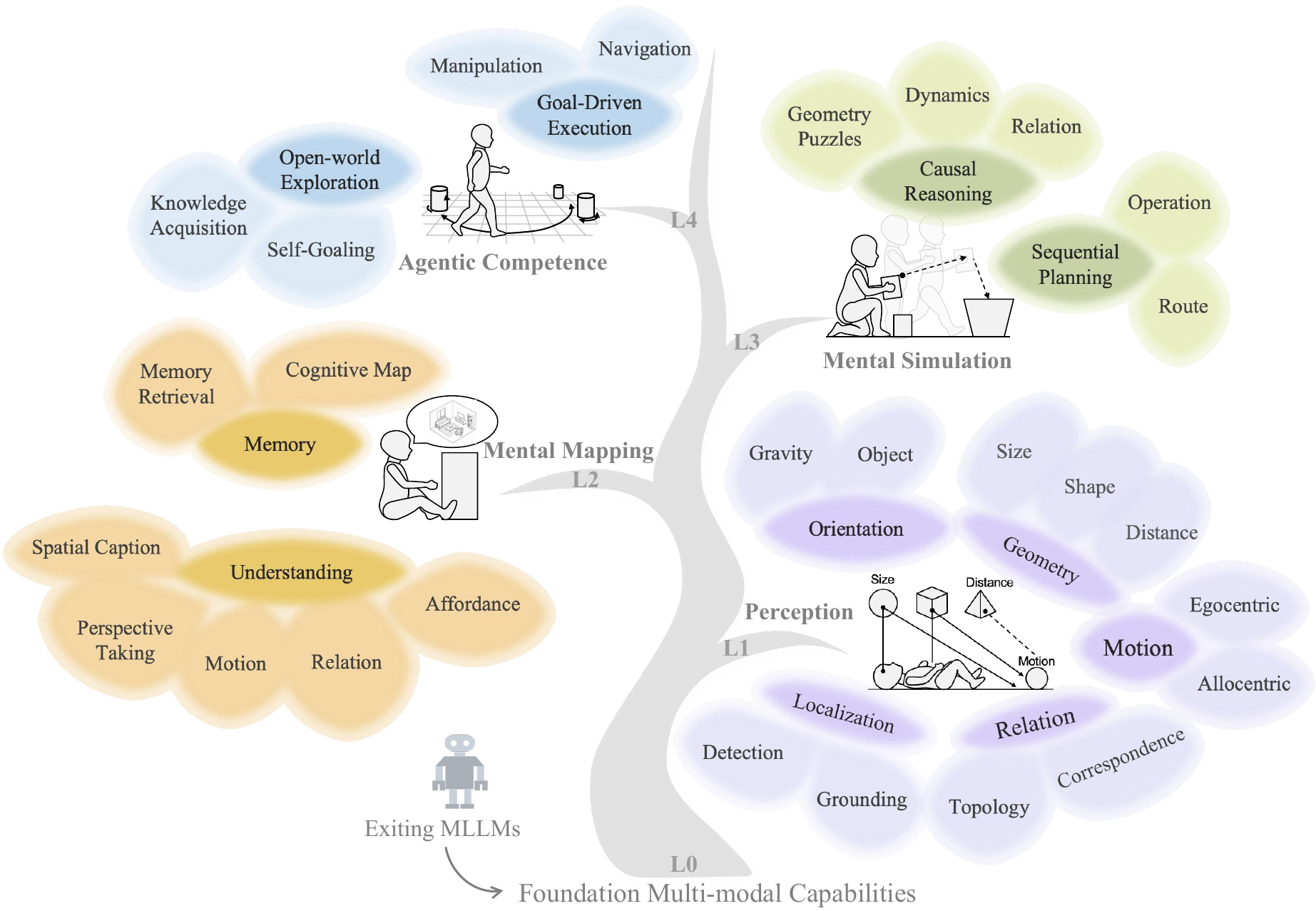}
\caption{\textbf{SpatialTree.} Inspired by cognitive science, our proposed SpatialTree organizes spatial intelligence into a four-layer hierarchy (L1-L4). Rooted in foundational multi-modal capabilities (L0), the tree progressively branches from Basic perception (L1) to agentic competence (L4).}
\label{fig:spatree}
\end{figure}

\footnotetext[1]{Work done during Yuxi Xiao and Longfei Li’s internship at Bytedance Seed.}

\section{Introduction}
\label{sec:intro}
Spatial abilities refer to the capacity to perceive, understand, reason about, and interact with 2D and 3D space, a long-standing topic in cognitive science~\citep{tolman1948cognitive, shepard1988mental, hegarty2010components}. In multimodal large language models (MLLMs), these abilities form the cornerstone of Spatial Intelligence (SI), yet remain challenging to study systematically due to their inherent complexity and broad scope~\citep{liu2025spatial, vst_yang2025}.

Recent research followed a task-centric trajectory. Early efforts concentrated on spatial tasks within single images~\citep{ma20243dsrbench, wang2024spatial,fu2024blink}, such as relative object positioning and size estimation. As capabilities matured, studies expanded into the 3D domain, tackling grounding, detection, and captioning from point clouds~\citep{zhu2024llava,hong20233d,xu2024pointllm}. More recently, the advent of multi-view and video-capable models has further diversified the landscape~\citep{yang2025thinking, wang2025site, wang2025spatialviz, gholami2025spatial, yang2025mmsi, jia2025omnispatial}, covering a broad spectrum from spatial relevant reasoning to egocentric and dynamic object understanding.

However, these task-centric benchmarks remain fragmented and often treat spatial capabilities as isolated or overlapping skills. This lack of unification makes it difficult to unravel the inherent structure and cross-level dependencies within the proliferation of spatial tasks. 

\begin{quote}
Can we move beyond disjointed, task-centric evaluations to uncover a compact set of atomic capabilities that reveal how spatial abilities emerge, interact, and transfer?
\end{quote}

Drawing inspiration from cognitive science insight that "intelligence is a dynamic structure built through successive stages"~\cite{piaget2013construction}, we advocate a paradigm shift: moving from fragmented tasks to a \emph{capability-centric} framework. Specifically, we structure Spatial Abilities as a four level capability tree (Fig.~\ref{fig:spatree}):

\begin{itemize}
    \setlength{\itemsep}{2pt}
    \item \textbf{L1 Perception:}
    This level focuses on native perception of space, capturing raw geometric and physical attributes such as size, distance, and motion, without relying on language or symbolic reasoning.

    \item \textbf{L2 Mental Mapping:}
    This level maps spatial perception to language, grounding spatial concepts in linguistic semantics and forming language-structured spatial memory.

    \item \textbf{L3 Mental Simulation:}
    This level supports internal reasoning about space, enabling mental simulation, including causal reasoning about dynamics, relational and geometric problem solving, and sequential planning for actions and navigation.

    \item \textbf{L4 Spatial Agent:}
    This level executes actions in space, integrating perception, language, and reasoning to interact with the environment, interpret feedback, and complete long-horizon spatial tasks.
\end{itemize}

To populate this taxonomy, we first reorganize and unify data from prior works~\citep{yang2025mmsi,yang2025thinking,jia2025omnispatial,lin2025towards,wang2025spatialviz,wang2024spatial,zhu2024llava,yin2025spatial,xu2025multi,liu2025ir3d}, which primarily cover specific sub-domains within L1 to L3. To address missing abilities and enrich annotations beyond existing datasets, we construct a Spatial Engine that integrates multiple expert models~\cite{yang2024depth,orient_anything,xiao2025spatialtrackerv2,mao2025spatiallm,veicht2024geocalib,li2025infinityhuman,xiao2024spatialtracker,liu2025trace}. For L1 to L3, we enhance data diversity by introducing multiple QA formats for the same underlying problem and by incorporating additional sub-capabilities, including orientation estimation, localization, and affordance understanding. In contrast, L4 requires agentic interactions that are largely absent from prior benchmarks. We therefore meticulously curate L4 data across three representative embodiments: character navigation, robot grippers, and human hands. To facilitate MLLM adaptation to agentic tasks, we design an action mapping strategy that discretizes low-level actions into high-level motion primitives, forming an executable action space. In addition, human–object interaction sequences are reformatted into multi-step multiple-choice tasks through manual annotation. Based on the resulting datasets and task formulations, we establish \textbf{SpatialTree-bench} to evaluate a broad range of mainstream open-source and commercial MLLMs.

Our evaluation on \textbf{SpatialTree-Bench} uncovers a clear hierarchical dependency: while foundational abilities (L1) function independently, higher-level capabilities exhibit strong reliance on these basics. To systematically validate this structure, we move beyond static evaluation to targeted training interventions using both Supervised Fine-Tuning (SFT) and Reinforcement Learning (RL).
First, our SFT experiments reveal that spatial skills do not scale uniformly. We observe that while basic abilities struggle with intra-level interference, they serve as critical \textit{stepping stones} for higher-level tasks, unlocking a powerful \textbf{multi-ability synergy} when jointly trained. Second, we explore the role of reasoning via RL, leading to a surprising dichotomy between ``thinking'' and ``perceiving''. We find that while extensive reasoning is indispensable for complex tasks, it can be detrimental to intuitive perception (e.g., numerical estimation), where over-thinking degrades precision. This necessitates a \textbf{auto-think strategy}: suppressing reasoning for immediate perception while encouraging it for complex planning.
In summary, our key contributions are:
\begin{itemize}
    \item We construct \textbf{SpatialTree}, the first capability-centric benchmark that organizes spatial intelligence into a rigorous hierarchy, enabling granular diagnosis of model capabilities.
    \item We reveal the structural transfer dynamics of spatial skills, demonstrating how foundational perception acts as a prerequisite for agentic competence through systematic SFT experiments.
    \item We identify the trade-off between reasoning and perception in RL training, proposing a differentiated inference strategy that balances intuitive sensing with complex reasoning.
\end{itemize}
\section{Related Work}
\label{sec:related_work}

\paragraph{\textbf{Spatial Cognitive Modeling.}} 
Understanding spatial cognition has long been a central goal in cognitive science and AI. A common insight from classical theories is that spatial abilities are hierarchical, ranging from basic perception and sensorimotor interactions to higher-level reasoning and planning. Piaget~\cite{piaget2013construction} highlighted the developmental progression of such abilities, Tolman~\cite{tolman1948cognitive} introduced the idea of cognitive maps to represent environments for flexible navigation, and Kuipers~\cite{kuipers1978modeling, kuipers2000spatial} formalized a hierarchical spatial representation linking local perception to global knowledge. More recent symbolic and neural approaches~\cite{shepard1988mental, newcombe2000making} extend these insights to computational models of spatial representation, memory, and reasoning. These studies collectively motivate our \emph{SpatialTree}, which organizes spatial intelligence into multi-level capabilities, bridging classical theory with systematic computational evaluation.

\paragraph{\textbf{Multi-modal Large Language Models.}} 
The success of GPT-3~\cite{brown2020language} and GPT-3.5~\cite{openai-gpt-3.5} demonstrated the potential of large language models for complex linguistic understanding and reasoning. GPT-4V~\cite{openai2023gpt4v} extends GPT-4~\cite{achiam2023gpt} with visual inputs, enabling single-image understanding and basic spatial reasoning. Open-sourced models such as LLaVA~\cite{liu2023visual} and QwenVL~\cite{bai2023qwen} gradually added multi-image and video capabilities, supporting spatiotemporal reasoning. Reasoning-augmented LLMs, pioneered by OpenAI O1~\cite{jaech2024openai} and DeepSeek-R1~\cite{guo2025deepseek}, integrate chain-of-thought and reinforcement learning to enhance high-level inference. Building on these advances, GPT-4O~\cite{hurst2024gpt} and Gemini 2.5~\cite{comanici2025gemini} combine perception and reasoning to support complex, agentic decision-making. Collectively, these milestones progressively enable hierarchical spatial intelligence in MLLMs, motivating structured benchmarks and evaluation frameworks across low-level perception, intermediate reasoning, and high-level agentic competence.

\paragraph{\textbf{Benchmarks for Spatial Intelligence in MLLMs.}} 
Benchmarks for spatial abilities in MLLMs have evolved alongside the models themselves. Early efforts, such as BLINK~\cite{fu2024blink}, SpatialEval~\cite{wang2024spatial}, and 3DSR-Bench~\cite{ma20243dsrbench}, focused on evaluating spatial understanding tasks in single images, including distance estimation, relational question answering, and spatial captions. As MLLMs increasingly support multi-frame and video inputs, benchmarks such as VSI-Bench~\cite{yang2025thinking} and MMSI-Bench~\cite{yang2025mmsi} have emerged to evaluate spatial reasoning across multiple views and dynamic scenes. To further enrich task diversity and coverage, Omnispatial~\cite{jia2025omnispatial}, SITE~\cite{wang2025site}, and IR3D-Bench~\cite{liu2025ir3d} extend benchmarks to geometry puzzles, dynamic reasoning, and inverse rendering tasks. Built upon prior efforts, our SpatialTree benchmark systematically organizes spatial abilities into a hierarchical framework, providing the first thorough evaluation across different capabilities.
\begin{figure}[!t]
    \centering
    \begin{subfigure}[b]{0.43\linewidth} 
        \centering
        \includegraphics[width=\linewidth]{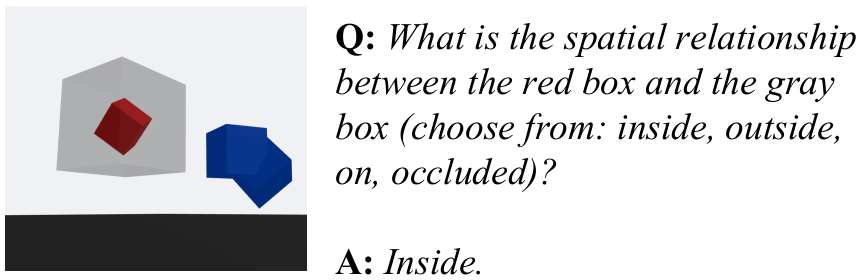}
        \caption{}
        \label{fig:sub_a}
    \end{subfigure}
    \begin{subfigure}[b]{0.56\linewidth} 
        \centering
        \includegraphics[width=\linewidth]{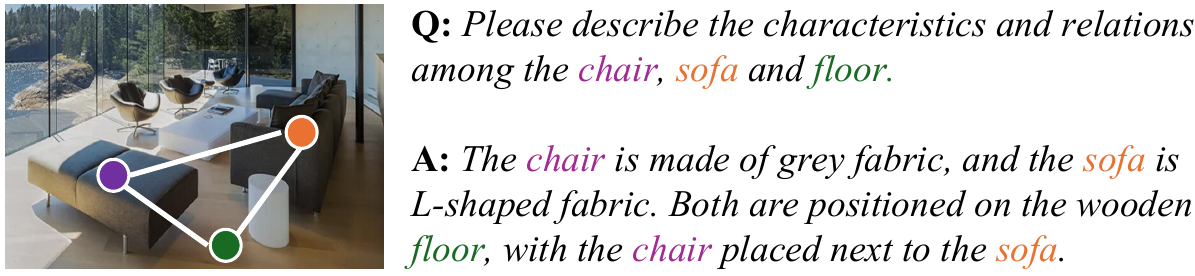}
        \caption{}
        \label{fig:sub_b}
    \end{subfigure}
    \vspace{1em} 
    \begin{subfigure}[b]{0.99\linewidth} 
        \centering
        \includegraphics[width=\linewidth]{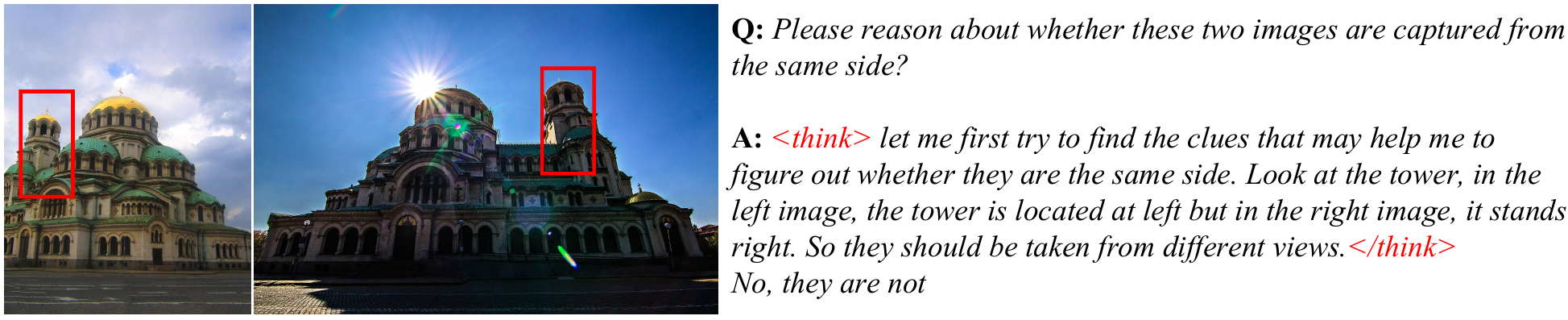}
        \caption{}
        \label{fig:sub_c}
    \end{subfigure}
    \caption{\textbf{Different Emphasis across Hierarchy Levels.} Taking \textit{Relation} in L1, L2, L3 as an example:
    \textbf{(a)} \textit{Relation} in \textit{Perception}, involving basic spatial relations (e.g., inside, outside)
    \textbf{(b)} \textit{Relation} in \textit{Understanding}, describing the attributes and mutual relationships among different objects.
    \textbf{(c)} \textit{Relation} in \textit{Causal Reasoning}, leveraging visual cues and logical inference to solve more complex relational tasks.}
    
    \label{fig:three_images}
\end{figure}

\section{The SpatialTree Taxonomy}
\label{sec:spatialtree_framework}

In this section, we introduce SpatialTree taxonomy of spatial capabilities. Different levels of the taxonomy emphasize different aspects of spatial ability, as illustrated in Fig.~\ref{fig:three_images}. Lower levels focus on intuitive and basic perceptual skills, while higher levels progressively require richer knowledge and more complex logical reasoning. Concrete instantiations are discussed in Sec.~\ref{sec:si-engine}.


\subsection{\textbf{Perception}}
\label{sec:l1-perception}
We begin with \textbf{Perception} (L1), which characterizes the most primitive capacity to sense visual signals from the environment. Rooted in the sensorimotor stage of human development, this level represents the instinctive ability to process spatial information prior to linguistic abstraction. Based on these fundamental biological needs, we categorize perception into five core abilities:

\textbf{Geometry} interprets the physical form and metric properties of the world. Evolutionarily, this allows agents to intuitively decipher three core dimensions: \textit{Distance} (gauging metric and relative depth), \textit{Size} (estimating physical magnitude, area, and volume), and \textit{Shape} (discerning contours, boundaries, and geometric primitives). These faculties enable immediate judgments, such as deciding if a fruit is small enough to hold or a path is wide enough to traverse.

\textbf{Motion} reflects the innate capacity to process dynamic visual signals over time. This perception is categorized into \textit{Egocentric} (sensing self-motion and heading direction) and \textit{Allocentric} (perceiving the movement and speed of external objects). Interpreting these signals allows humans to instantly determine if an entity is static or mobile, supporting tasks like avoiding obstacles or tracking targets.
    
\textbf{Orientation}  is rooted in the necessity to maintain balance and align with the environment. Humans instinctively sense ``up'' and ``down'' via vestibular cues to avoid falling, and visually recognize object poses to interact with them effectively. This category includes \textit{Gravity} (sensing the vertical axis, e.g., knowing which way is up) and \textit{Object} (perceiving poses, e.g., noticing a cup is tilted).
    
\textbf{Relation} captures the structural continuity and spatial arrangement of the visual world. Beyond isolating individual objects, intuitive perception registers how entities are spatially arranged and whether they correspond across views. This category includes \textit{Topology} (perceiving basic spatial configurations such as \textit{inside}, \textit{outside}, or \textit{overlap}) and \textit{Correspondence} (recognizing the same object or landmark across different viewpoints or visual conditions).
    
\textbf{Localization} anchors visual stimuli within 2D/3D space. It addresses the fundamental question of where, allowing humans to spot a friend in a crowd or locate keys on a cluttered table. This category includes \textit{Detection} (identifying object presence and spatial extent) and \textit{Grounding} (associating visual observations with spatial positions or coordinates).

\subsection{\textbf{Mental Mapping}}
\label{sec:l2-mental-mapping}
\textbf{Mental Mapping} (L2) marks a shift toward alignment with language. It encompasses two key aspects: mapping spatial primitives to semantic concepts, and constructing a language-aligned memory system. Accordingly, we classify Mental Mapping into two main sub-abilities:

\textbf{Understanding} interprets the semantic meaning underlying geometric perception, shifting the focus from mere existence to function and identity. This capability begins by translating visual scenes into linguistic descriptions through \textit{Spatial Captioning}, and by identifying semantic \textit{Relations} (e.g., distinguishing “riding” from “sitting on”). It further elevates \textit{Motion} from raw kinematic cues to semantic interpretation, recognizing purposeful actions rather than simple movement. Crucially, this level includes \textit{Perspective Taking}—the ability to mentally align with alternative viewpoints—and \textit{Affordance} understanding, which identifies the functional possibilities of objects (e.g., recognizing that a handle is graspable), thereby bridging perception with potential future interactions.

\textbf{Memory} extends spatial awareness beyond the instantaneous field of view, enabling the system to retain and update spatial information over time. This capability relies on constructing a \textbf{Cognitive Map}, which synthesizes fragmentary observations (e.g., video frames or multi-view images) into a compact and unified global representation. Based on this mental model, the system performs \textbf{Memory Retrieval} to ground specific semantic events or moments, allowing it to recall where an object appeared or when a specific action occurred, even if the target is currently occluded or out of sight.


\subsection{\textbf{Mental Simulation}}
\label{sec:l3-mental-simulation}
Reasoning and planning prior to action execution are essential components of MLLMs, aligning naturally with the Chain-of-Thought paradigm in language model reasoning. In spatial cognitive science, this process is commonly referred to as mental simulation. We taxonomize mental simulation into two orthogonal directions: causal reasoning and sequential planning.

\textbf{Causal Reasoning} allows MLLMs to model spatial interactions, physical dynamics, and entity relationships within a simulated mental space. It includes reasoning about object geometry (e.g., how shapes interlock in spatial puzzles), motion prediction (e.g., how an object traverses a path), and analyzing semantic–spatial relations (e.g., object A is left of object B). By mentally simulating cause–effect chains in spatial scenarios, MLLMs establish the logical substrate for subsequent planning.

\textbf{Sequential Planning} converts causal insights into coherent, goal-directed action plans expressed in language. It entails designing high-level, step-by-step strategies (e.g., "first move toward the door, then turn right, and finally interact with the handle") and generating abstract routes that respect spatial logic (e.g., "go around the table to reach the sofa"). By chaining linguistic action primitives, MLLMs produce strategic plans that ensure the conceptual sequence aligns with the overarching goal before any low-level execution.

\subsection{\textbf{Agentic Competence}}
\label{sec:l4-agentic-competence}
Agentic Competence represents the culmination of spatial intelligence, bridging the gap between cognitive planning and practical execution. It focuses on the agent's ability to translate internal plans into tangible interactions within dynamic environments. This capability is grounded in diverse embodied scenarios, where the model must interpret visual streams to generate precise action sequences. Key tasks include predicting control commands from \textbf{video game frames}, deriving manipulation sequences from \textbf{robotic arm videos}, and identifying potential movement directions or navigational affordances in \textbf{real-world scenes}.

We begin from the ultimate objective of a \textit{Spatial AI Agent} --- an MLLM-driven system that integrates multi-modal observations, updates its memory, and selects actions to interact with the 3D world in an intuitive manner. Formally, the agent performs sequential decision-making by modeling:
\begin{align}
(S_t, A_t, M_t) & \sim P_\theta \Big( \cdot \;\big|\; O_t, H_{t-1} \Big), \label{eq:agent_model}
\text{where} \quad H_{t-1}  = \left\{ (O_0, A_0, M_0), \dots, (O_{t-1}, A_{t-1}, M_{t-1}) \right\} \nonumber
\end{align}
\vspace{-0.2em}
where $O_t \in \mathcal{O}$ is the current multi-modal observation, $S_t \in \mathcal{S}$ the internal latent state (e.g., goal, plan, or belief), $A_t \in \mathcal{A}$ the chosen action, and $M_t \in \mathcal{M}$ the updated memory representation. MLLMs are expected to output interactive actions executable across 3D environments and embodiments, such as games, simulators, and the physical world. Unlike Vision-Language Action Models (VLAs) decording the low-level control signals in robotics~\citep{intelligence2025pi_}, MLLMs take the language as the only interface to link with environments like GUI Agents~\citep{UI-TARS}.  

    

\begin{figure}[!t]
    \centering
    \includegraphics[width=\linewidth]{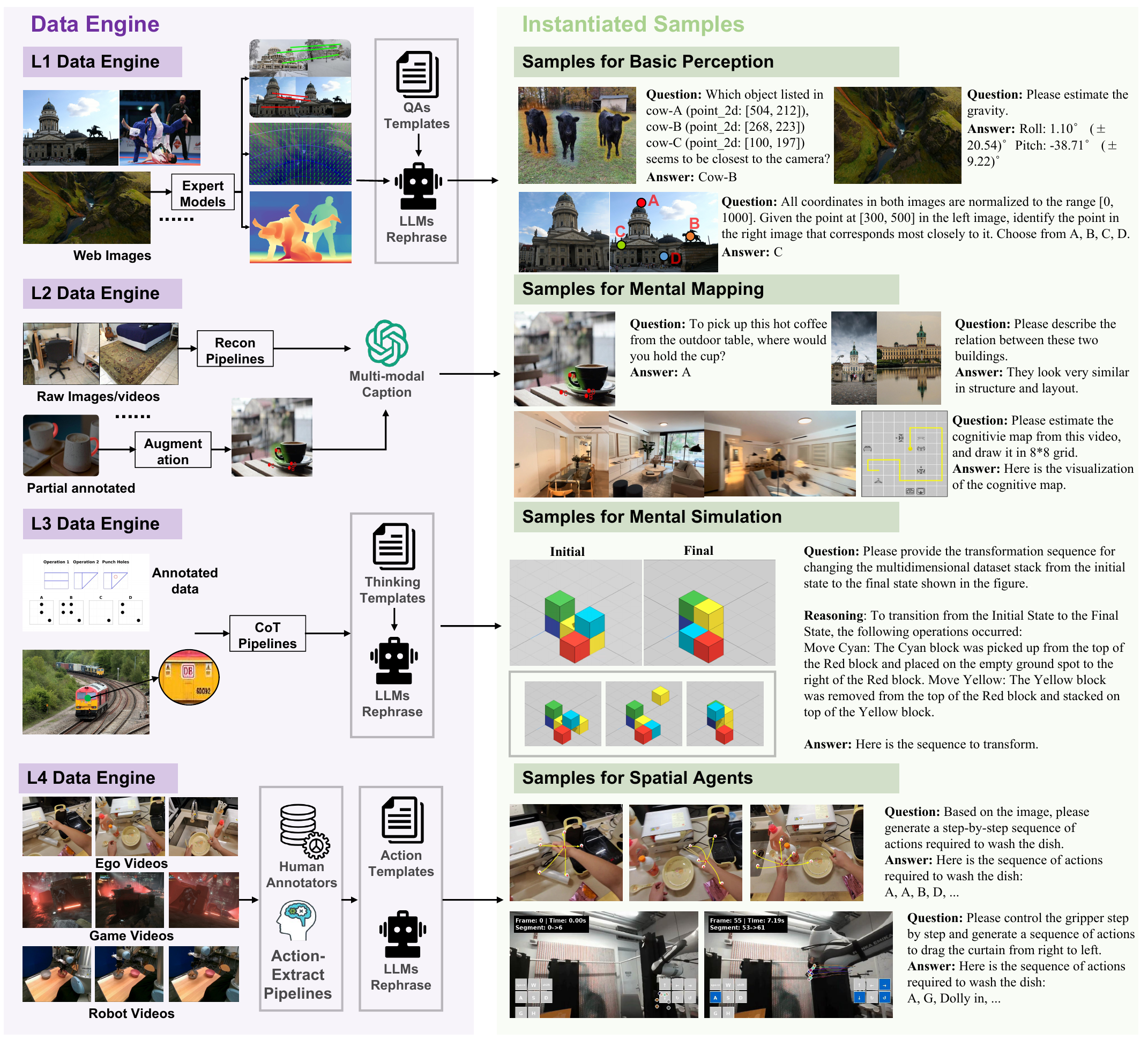}
    \caption{\textbf{Benchmark Data Engines.} Level-specific engines process data and construct QAs.}
    \label{fig:bench_engine}
\end{figure}

\section{Instantiating the SpatialTree Benchmark}
\label{sec:si-engine}
This section details how the SpatialTree taxonomy is instantiated as a benchmark, covering data curation pipeline (seen in in Fig.~\ref{fig:bench_engine}), and more specific examples of the QAs are demonstrated in Sec.~\ref{sec:vis_qa}. 


\subsection{Data Curation}

\textbf{Data Annotation Framework.}
As shown in Fig.~\ref{fig:bench_engine}, the data engines are organized hierarchically. At the perception level, we employ a set of expert perception models, including DepthAnything3~\citep{yang2024depth}, SpatialTracker~\citep{xiao2025spatialtrackerv2,xiao2024spatialtracker}, GeoCalib~\citep{veicht2024geocalib}, and OrientAnything~\citep{orient_anything}, to extract intermediate perceptual representations, such as depth, correspondences, tracking results, and gravity direction. These representations are then further processed using our designed QA templates and rephrased by LLMs to construct the target QAs. For L2, we leverage 3D reconstruction pipelines (Sec.~\ref{appsubsec:agentic_com}) to generate BEV maps from videos, which are subsequently captioned and transformed into QAs for cognitive mapping and memory retrieval with multimodal LLMs. 
For other understanding tasks, such as affordance, we build on partially annotated datasets~\citep{luo2022learning} and augment them by randomly constructing multiple-choice candidates through various visual prompts and by reformulating questions with more abstract descriptions. For L3, we build upon annotated reasoning-related QAs and introduce structured thinking templates, enabling an LLM-based rephraser to augment the original QAs with explicit Chain-of-Thought (CoT) reasoning. For agentic tasks, we curate Internet data covering manipulation and navigation across diverse embodiments (e.g., human hands, robotic grippers, and game characters). We further process the data with our Action-Extraction Pipeline (Sec.~\ref{appsubsec:agentic_com}), converting embodiment-specific actions into unified key–mouse action sequences suitable for MLLMs. In parallel, we introduce a human annotation pipeline to annotate videos with executable multi-option action sequences.

\textbf{Data Resources.}
Our SpatialTree-Bench is constructed by systematically reorganizing numerous recent datasets (detailed in the Appendix)~\citep{yang2025thinking, yang2025mmsi, zhu2024llava, wang2024spatial, yin2025spatial, lin2025towards, jia2025omnispatial, yang2025embodiedbench, wang2025spatialviz, xu2025multi, ma20243dsrbench} to address their scattered capability coverage and over-reliance on simple questions. We first map each question to our SpatialTree framework and then enhance the evaluation protocol; for instance, complex reasoning tasks from CameraBench and MMSI-Bench are converted to a hybrid \textit{multi-option + LLM-as-a-Judge} format for a finer-grained assessment. To fill the remaining capability gaps, we introduce \textbf{SpatialPlus}, a new dataset targeting underrepresented abilities (e.g., L1 Orientation, L1 Shape, L2 Spatial Caption) with a primary emphasis on L4 Agentic Competence. We leverage our proprietary \textbf{SpatialEngine} to automatically create annotations from a diverse array of video sources, including 3D reconstruction datasets, in-game footage~\cite{ju2024miradata}, egocentric manipulation videos~\citep{egodex}, and robotics data~\citep{khazatsky2024droid}. More implementation details are discussed in Appendix.

\subsection{Evaluation Metrics Designs}
\begin{figure}[!t]
    \centering
    \includegraphics[width=\linewidth]{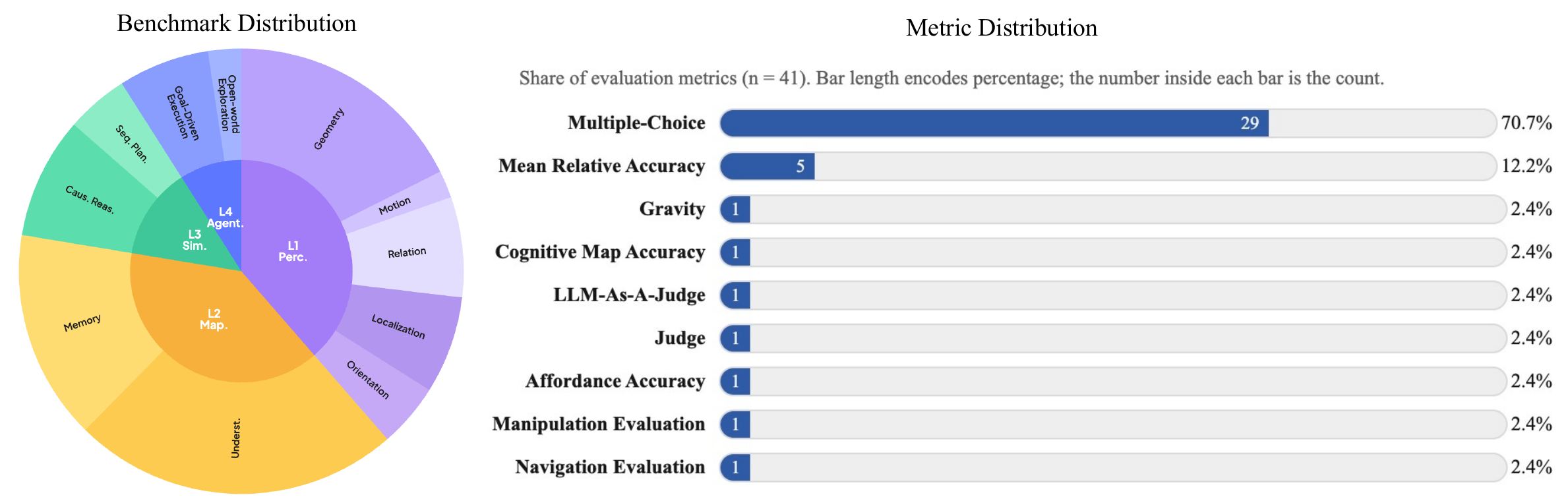}
    \caption{\textbf{Distribution of Benchmark data and Evaluation Metrics.} We analyze the metric usage across 41 tasks in our benchmark. The evaluation relies primarily on multiple-choice questions (70.7\%), complemented by task-specific numeric metrics (e.g., cognitive map accuracy) and LLM-as-a-Judge protocols.}
    \label{fig:data_metric_dist}
\end{figure}

To facilitate robust evaluation within our hierarchical benchmark, we design diverse metrics tailored to specific downstream tasks. As illustrated in Fig.~\ref{fig:data_metric_dist}, these metrics are primarily categorized into multiple-choice questions (70.7\%), numeric accuracy (e.g., mean relative accuracy), LLM-based evaluation (LLM-as-a-Judge) and specific numeric metrics.

\section{A Hierarchical Analysis of Spatial Capabilities}
\label{sec:evaluation_benchmark}

\subsection{Models and Metrics}
\label{sec:models_and_metrics}

\textbf{Benchmarked Models.}
We categorize the evaluated MLLMs into three groups:
(1) Thinking Models, i.e., models augmented with explicit reasoning or chain-of-thought generation mechanisms (reasoning-augmented), including Gemini 3 Pro~\citep{gemini3pro2026}, Gemini 3 Flash~\citep{gemini3flash2026}, Gemini 2.5 Flash, Gemini 2.5 Pro~\citep{comanici2025gemini}, GLM-4.5V~\citep{hong2025glm}, Seed1.5VL~\citep{guo2025seed1} and Seed1.8~\citep{seedseed1};
(2) Non-Thinking Models, which do not explicitly optimize for reasoning-style generation, such as Gemini 2.5-Pro-Nonthink, Gemini 2.5-Flash-Nonthinking, and GPT-4o~\citep{hurst2024gpt}; and
(3) Open-Source Models, including Qwen2.5-VL~\citep{bai2025qwen2}, Qwen3-VL~\citep{yang2025qwen3}, and Kimi-VL~\citep{team2025kimi}, representing recent community-driven multimodal advances. This diverse selection enables comprehensive comparisons across reasoning and non-reasoning paradigms, proprietary and open-source ecosystems, and model scales ranging from 32B to 72B parameters—providing a holistic overview of the current MLLM landscape. A full list of evaluated models is shown in Tab.~\ref{tab:main_table}.

\textbf{Evaluation Metrics.}
Our evaluation employs a multi-faceted set of metrics tailored to the specific abilities at each level of the SpatialTree. For perception and understanding tasks (L1-L2), we primarily use accuracy-based metrics, such as classification accuracy for object recognition, Mean Squared Error (MSE) for distance estimation, and angular difference for orientation tasks. For higher-level reasoning and planning tasks (L3-L4), we measure task success rates. In the case of agentic tasks (L4), we further analyze the quality of generated actions using metrics like positional error (L2 distance) and orientation error (angular difference) against ground-truth trajectories.
\subsection{Overall Performance}
\label{sec:spatialtree_bench}
\begin{table*}[ht!]
    \vspace{-0.4cm}
    \centering
    \resizebox{\linewidth}{!}{
    \begin{tabular}{l|cc|ccccc|cc|cc|cc}
    
    \hline 
    
    \rowcolor{white} 
    & & & 
    \multicolumn{5}{c|}{\cellcolor{L1_Purple} L1 Perception} &  
    \multicolumn{2}{c|}{\cellcolor{L2_Orange} L2 Mental Mapping} &
    \multicolumn{2}{c|}{\cellcolor{L3_Green} L3 Mental Simulation} &
    \multicolumn{2}{c}{\cellcolor{L4_Blue} L4 Agentic Competence} \\
    
    \rowcolor{white} 
    \textbf{Methods} & \textbf{Rank} & \textbf{Avg.} & 
    \textbf{Geom.} & \textbf{Motion} & \textbf{Rel.} & \textbf{Local.} & \textbf{Orient.} &
    \textbf{Underst.} & \textbf{Memory} &
    \textbf{Caus. Reas.} & \textbf{Seq. Plan.} &
    \textbf{Goal Exec.} & \textbf{Open Expl.} \\
    
    \rowcolor{white}
    & & & 
    \small\textcolor{gray}{[0.40]} & \small\textcolor{gray}{[0.15]} & \small\textcolor{gray}{[0.15]} & \small\textcolor{gray}{[0.20]} & \small\textcolor{gray}{[0.10]} &
    \small\textcolor{gray}{[0.70]} & \small\textcolor{gray}{[0.30]} &
    \small\textcolor{gray}{[0.65]} & \small\textcolor{gray}{[0.35]} &
    \small\textcolor{gray}{[0.50]} & \small\textcolor{gray}{[0.50]} \\

    \hline

    \rowcolor{navyblue!5}
    \multicolumn{14}{l}{\textcolor{black}{\textit{Non-Thinking Models}}} \\
    GPT-4o & 13 & 31.9 & 23.9 &  38.6 & 29.8 & 24.2 & 36.2 & 31.2 & 43.6 & 29.3 & 40.5 & 25.8 & 39.2 \\ 
    Gemini2.5 Flash NT & 10 & 35.8 & 31.6 & 29.3 & 30.8 & 35.2 & 45.4 & 36.4 & 53.7 & 28.4 & 36.9 & 27.6 & 45.7 \\
    Gemini2.5 Pro NT & 5  & 41.4 & 36.2 & 30.0 & 33.2 & 47.0 &  48.5 & 43.3 & 55.2 & 39.6 & 47.5 &  29.2 & 46.0 \\
    
    \hline

    \rowcolor{navyblue!5}
    \multicolumn{14}{l}{\textcolor{black}{\textit{Thinking Models}}} \\
    Seed1.5vl  & 6 & 41.3 & 39.2 & 36.6 & 24.6 & 44.7 & 44.8  & 46.5 & 38.5  & 36.6 & 47.0 & 28.4 & 25.9 \\
    Seed1.8  & \cellcolor{oai-green-200}{3} & 50.3 & 42.5 & 42.7 & 49.9 & 54.5 & 52.4  & 56.3 & 51.0  & 49.7 & 53.6 & 26.0 & 70.6 \\
    GLM4.5V & 9  & 36.0 & 35.3 & 24.0 & 32.5 & 34.5 & 43.7 & 34.5 & 34.1 & 33.8 & 41.0 & 26.8 & 49.7 \\ 
    Gemini2.5-Pro & 4  & 50.1 & 47.8 & 32.6 & 44.4 & 61.6 & 47.9 & 50.5 & 61.5 & 47.6 & 58.2 & 28.3 &  63.3 \\
    Gemini2.5-Flash & 7 & 41.1 & 35.6 & 29.3 & 40.5 & 56.2 & 44.6 & 43.1 & 51.4 & 36.4 & 49.1 & 26.9 & 46.6 \\ 
    Gemini3-Pro & \cellcolor{oai-green-400}{2}  & 56.5 & \cellcolor{oai-gray-600}{\textcolor{white}{54.5}} & \cellcolor{oai-gray-600}{\textcolor{white}{47.4}} & 62.5 & \cellcolor{oai-gray-600}{\textcolor{white}{74.0}} & 53.9 & 60.5 & 50.3 & 56.0 & 68.2 & 29.9 &  68.5 \\
    Gemini3-Flash & \cellcolor{oai-green-600}{1}  & \cellcolor{oai-gray-600}{\textcolor{white}{57.8}} & 50.1 & 40.7 & \cellcolor{oai-gray-600}{\textcolor{white}{62.9}} & 62.4 & \cellcolor{oai-gray-600}{\textcolor{white}{54.6}} & \cellcolor{oai-gray-600}{\textcolor{white}{62.0}} & \cellcolor{oai-gray-600}{\textcolor{white}{66.8}} & \cellcolor{oai-gray-600}{\textcolor{white}{58.4}} & \cellcolor{oai-gray-600}{\textcolor{white}{68.6}} & \cellcolor{oai-gray-600}{\textcolor{white}{31.6}} &  \cellcolor{oai-gray-600}{\textcolor{white}{70.8}} \\
    \hline

    \rowcolor{navyblue!5}
    \multicolumn{14}{l}{\textcolor{black}{\textit{Open-source Models}}} \\

    Qwen2.5VL-7B & 15 & 27.5  & 17.8 & 22.6 & 23.9 & 20.6 & 31.6 & 34.7 & 15.2 & 28.4 & 39.8 & 24.5 & 31.1 \\
    Qwen2.5VL-32B & 14  & 27.9 & 21.4 & \cellcolor{oai-gray-300}{30.0} & 25.8 & 22.0 & 35.1 & 29.0 & 21.7 & 32.8 & 37.0 & 14.1 & 38.6 \\
    Qwen2.5VL-72B & 12  & 33.0 & 24.4 & 22.0 & 28.6 & \cellcolor{oai-gray-300}{37.5} & 37.3 & 38.9 & 35.1 & 32.6 & 38.4 & 23.9 & 38.6 \\
    Qwen3VL-30B & 11  & 35.3 & 31.9 & 26.0 & 32.3 & 20.7 & \cellcolor{oai-gray-300}{39.2} & 37.8 & 48.1 & 32.7 & 44.2 & 25.8 & 40.9 \\
    Qwen3VL-235B & 8  & \cellcolor{oai-gray-300}{40.0} & \cellcolor{oai-gray-300}{33.9} & 27.4 & \cellcolor{oai-gray-300}{35.1} & 35.4 & 38.9 & \cellcolor{oai-gray-300}{43.7} & \cellcolor{oai-gray-300}{53.4} & \cellcolor{oai-gray-300}{37.3} & \cellcolor{oai-gray-300}{44.7} & \cellcolor{oai-gray-300}{28.8} & \cellcolor{oai-gray-300}{48.9} \\ 
    Kimi-VL-A3B & 16 & 24.4 & 13.8 & 23.3 & 31.6 & 27.0 & 21.4 & 24.9 & 28.3 & 26.9 & 27.8 & 15.7 & 32.6 \\
    \hline
    \end{tabular}
    } 
    \vspace{-0.2cm}
    \caption{\textbf{Our-Bench.} \colorbox{oai-gray-600}{\textcolor{white}{Dark gray}} indicates the best result among all models and \colorbox{oai-gray-300}{light gray} indicates the best result among open-source models. NT denotes the non-thinking model. \textbf{Avg} is computed using the weights in brackets [\textcolor{gray}{$\cdot$}].}
    \label{tab:main_table}
\end{table*}

We first present the overall performance of all benchmarked models on our proposed SpatialTree-Bench, with detailed results summarized in Tab.~\ref{tab:main_table}. In our benchmark, Gemini 3 Flash achieves the best results (57.8) and Qwen3VL-235B get 40.0 leads the benchmark among open-source models.


\section{Exploring Ability Dependencies and Hierarchical Transfer}
\label{sec:ability_influence}

\FloatBarrier
\begin{wrapfigure}{R}{0.55\textwidth}
  \centering
  \vspace{-10pt} 
  \includegraphics[width=0.55\textwidth]{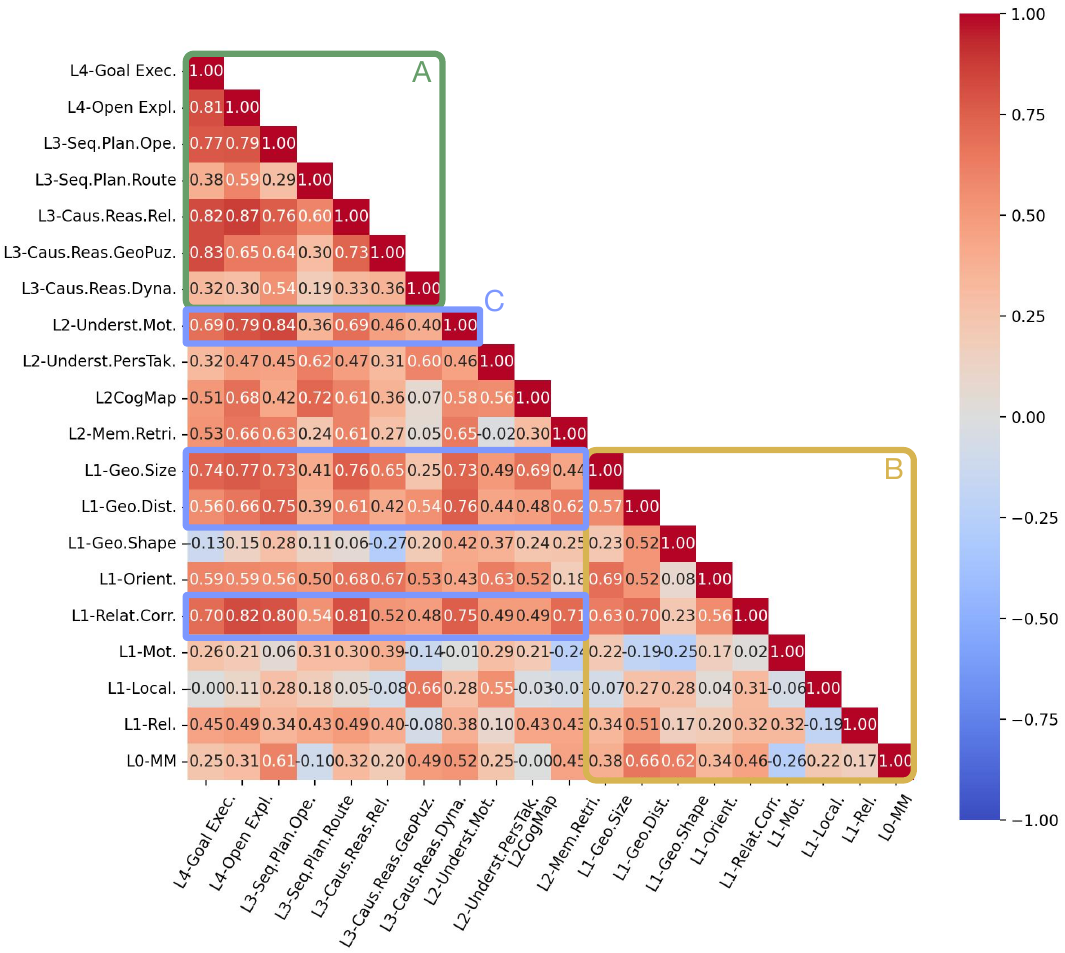}
  \caption{\textbf{Inter-Capability Dependencies via Pearson Correlation.} (A) Correlation matrix among higher-level capabilities (L3 and L4); (B) Correlation matrix among foundational L1 capabilities; (C) Salient low-level abilities influencing higher-level tasks.}
  \label{fig:wrap_example}
  \vspace{-20pt} 
\end{wrapfigure}

To investigate the structure of spatial ability in MLLMs, we analyze dependencies among fine-grained sub-abilities using Pearson correlation coefficients computed from our benchmark scores. A high positive correlation indicates that strong performance on one ability tends to accompany strong performance on another. Fig.~\ref{fig:wrap_example} presents a heatmap of these correlations across all models. Notably, higher-level capabilities (L3 and L4) exhibit stronger correlations (region A), suggesting that complex tasks such as route planning and causal reasoning rely on overlapping foundational sub-skills. In contrast, lower-level abilities (L1) show weak correlations, indicating they are largely independent. Based on this coarse correlation analysis, we select several underutilized low-level abilities for further investigation. These abilities are explored through Supervised Fine-Tuning (SFT) and explicit prompting to examine their influence and transfer, both within the same level and across higher levels.


\subsection{Probing Cross-Ability Transfer via SFT}
\label{sec:sft_exp}
\begin{table*}[ht!]
    \centering
    \resizebox{\linewidth}{!}{
    \begin{tabular}{l|c|ccccc|cc|cc|cc}
    
    \hline 
    
    \rowcolor{white} 
     & & 
    \multicolumn{5}{c|}{\cellcolor{L1_Purple} L1 Perception} &  
    \multicolumn{2}{c|}{\cellcolor{L2_Orange} L2 Mental Mapping} &
    \multicolumn{2}{c|}{\cellcolor{L3_Green} L3 Mental Simulation} &
    \multicolumn{2}{c}{\cellcolor{L4_Blue} L4 Agentic Competence} \\
    
    \rowcolor{white} 
    \textbf{Methods} & \textbf{Avg.} & 
    \textbf{Geom.} & \textbf{Motion} & \textbf{Rel.} & \textbf{Local.} & \textbf{Orient.} &
    \textbf{Underst.} & \textbf{Memory} &
    \textbf{Caus. Reas.} & \textbf{Seq. Plan.} &
    \textbf{Goal Exec.} & \textbf{Open Expl.} \\

    \hline

    \rowcolor{navyblue!5}
    Baseline & 25.0 & 20.9 & 28.6 & 28.9 & 24.2 & 34.2 & 22.6 & 21.7 & 27.2 & 31.7 & 22.1 & 26.5 \\ 
    B+Dist. & 24.5 & \goodres{24.1}{3.2} & \badres{26.6}{-2.0} & \badres{23.2}{-5.8} & \badres{19.6}{-4.6} & \neutralres{34.3}{0.1} & \goodres{24.6}{2.0} & \neutralres{21.8}{0.1} & \neutralres{26.1}{-1.1} & \neutralres{30.8}{-0.9} & \goodres{25.5}{3.4} & \neutralres{26.1}{-0.4} \\ 
    B+Corr. & 25.2 & \badres{17.6}{3.2} & \badres{23.9}{-4.7} & \goodres{30.2}{1.3} & \badres{18.9}{-5.3} & \goodres{35.6}{1.4} & \neutralres{21.9}{-0.7} & \goodres{24.6}{2.9} & \badres{21.8}{-5.4} & \goodres{33.9}{2.2} & \goodres{24.7}{2.6} & \goodres{35.9}{9.4} \\ 
    B+Size & 23.5 & \goodres{24.3}{3.4} & \badres{22.6}{-6.0} & \badres{21.4}{-7.5} & \badres{21.7}{-2.5} & \neutralres{34.5}{0.3} & \neutralres{21.9}{-0.8} & \badres{19.2}{-2.5} & \badres{23.4}{-3.8} & \badres{30.3}{-1.5} & \badres{21.5}{-0.6} & \badres{24.3}{2.2} \\ 
    B+Dist.+Size+Corr. & 26.1 & \goodres{25.5}{4.6} & \neutralres{29.3}{0.7} & \neutralres{29.4}{0.5} & \badres{16.4}{-7.8} & \neutralres{33.7}{0.5} & \goodres{23.0}{0.4} & \goodres{24.2}{2.5} & \badres{25.2}{-2.0} & \goodres{34.2}{2.5} & \goodres{26.0}{3.9} & \goodres{28.5}{2.0} \\ 
    B+Dist.+Size+Corr.+Mot. & 27.3 & \goodres{28.6}{7.7} & \badres{24.6}{-4.0} & \badres{20.6}{-8.3} & \goodres{26.3}{2.1} & \goodres{36.0}{1.8} & \neutralres{22.2}{-0.4} & \neutralres{22.6}{0.9} & \goodres{28.2}{1.0} & \goodres{32.8}{1.1} & \goodres{23.3}{1.1} & \goodres{35.9}{9.4} \\ 
    Baseline+75@(all spat.) & 23.6 & \goodres{24.9}{4.0} & \badres{22.6}{-6.0} & \badres{25.9}{-3.0} & \badres{17.4}{-6.8} & \badres{31.2}{-3.0} & \neutralres{22.2}{-0.4} & \badres{20.6}{-1.1} & \badres{25.7}{-1.5} & \badres{30.2}{-1.5} & \badres{19.7}{-2.4} & \badres{22.8}{-3.7} \\ 
    \hline
    \end{tabular}
    } 
    \caption{\textbf{SFT Comparisons.}
    "B+Dist.", "B+Corr.", and "B+Size" denote the baseline augmented with distance, correspondence, and size tuning data, respectively. Changes are color-coded as \textcolor{papergreen}{notable gains}, \textcolor{gray}{neutral influence}, and \textcolor{paperred}{drops}.}
    \label{tab:sft_exp}
\end{table*}

\begin{figure}[t]
\centering
\includegraphics[width=0.98\linewidth]{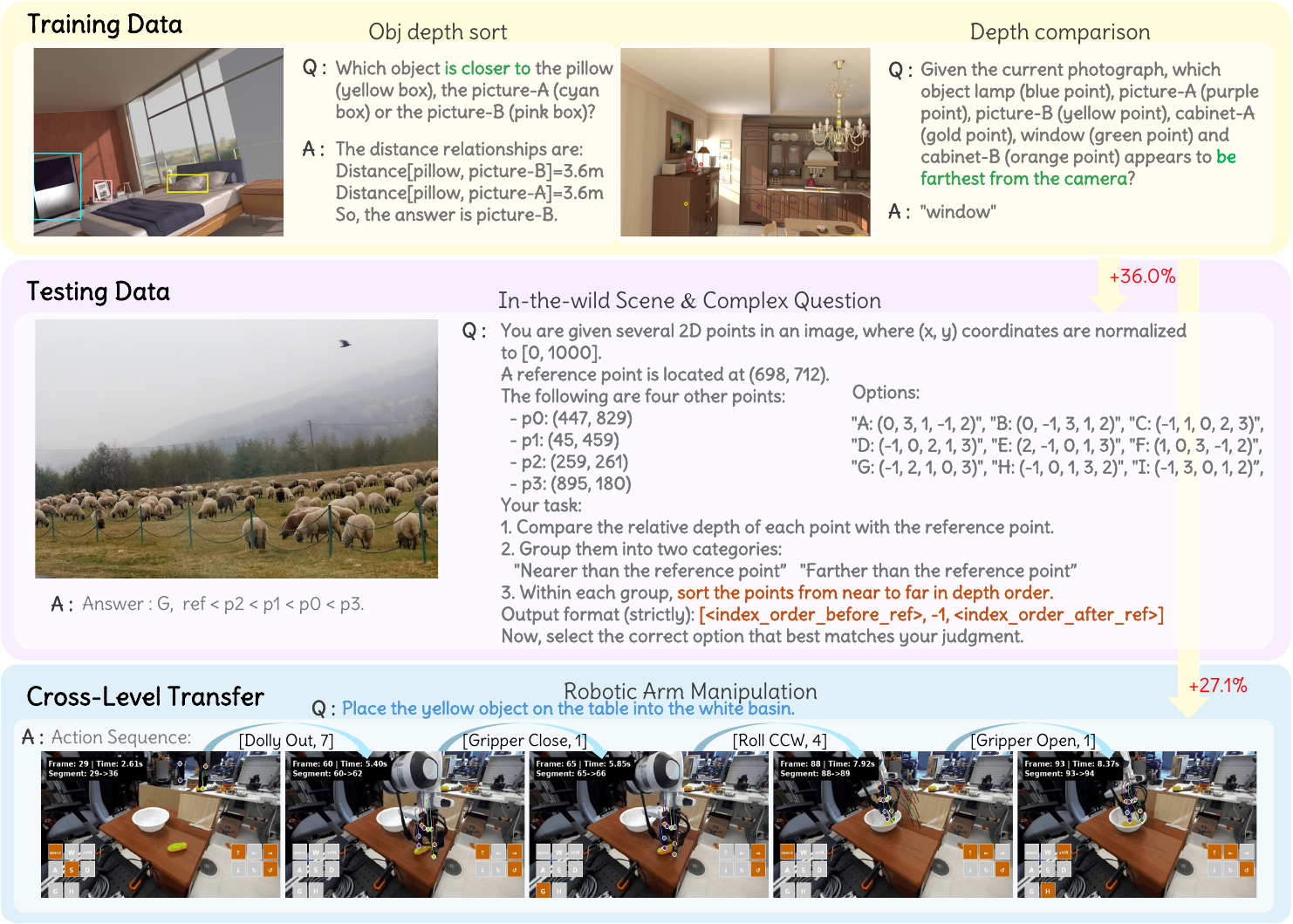}
\caption{\textbf{Demonstration of Capability Transfer after Distance SFT.} 
\textbf{(Top)} The model is trained on distance QAs, such as object depth sorting and comparison, just using data from synthetic and indoor scenes. 
\textbf{(Middle)} This learned capability transfers in a zero-shot manner to complex reasoning tasks in unseen, in-the-wild scenes, achieving a \textbf{36.0\%} performance gain over the baseline. 
\textbf{(Bottom)} Furthermore, the skill exhibits cross-level transfer, enabling the model to perform a robotic arm manipulation task with a \textbf{27.1\%} performance gain.}
\label{fig:ability_transfer}
\end{figure}

\begin{tcolorbox}[
    colback=seedblue!5,
    colframe=seedblue!80,
    boxrule=0.5pt,
    arc=2pt,
    left=6pt,
    right=6pt,
    top=6pt,
    bottom=6pt,
    fonttitle=\bfseries,
    title=\textbf{Finding 1}
]
\textbf{Cross-Ability Transfer:} Single-ability L1 SFT induces cross-level transfer, while yielding limited or slightly negative effects on same-level abilities.
\end{tcolorbox}

Based on a naive Pearson correlation analysis, we manually select three L1 abilities that exhibit the strongest correlations with higher-level performance: Geometry Distance (\textit{L1-Geo.Dist}), Geometry Size (\textit{L1-Geo.Size}), and Relative Correlation (\textit{L1-Relat.Corr}). 

\noindent\textbf{General Data Mixture.} To construct the general visual-instruction data, we follow the VST~\cite{vst_yang2025} data mixing recipe and combine multimodal datasets from LLaVA-Video~\citep{zhang2024llavanext-video}, LLaVA-NeXT-Interleave~\citep{li2024llava_next_interleave}, and LLaVA-OneVision~\citep{li2024llava_onevision}, covering single-image, multi-image, video, and 3D tasks. We then use SpatialEngine to generate ability-specific instruction data, mixed with the general data in a 1:3 ratio specifically for each. To isolate the gains from general data, we use a baseline fine-tuned only on the general data with the same token consumption. 

\noindent\textbf{Targeted SFT Data.} For \textit{L1-Geo.Dist}, we generate approximately 0.25M distance-relevant QA samples from SUNRGBD~\citep{song2015sun}, Hypersim~\citep{Hypersim}, and Matterport3D~\citep{Matterport3D}. The training data is further augmented with visual prompts and multi-scale transformations to enhance distance reasoning. For \textit{L1-Relat.Corr}, we generate matching data following VST~\cite{vst_yang2025}, sampling 0.25M examples. Similarly, for \textit{L1-Geo.Size}, we generate 0.25M samples from 3D bounding-box annotated datasets, including SUNRGBD, Hypersim and ArkitScenes~\citep{baruch2021arkitscenes}. 

\noindent\textbf{Results and Analysis.} As shown in Tab.~\ref{fig:ability_transfer}, single-ability SFT on distance, correspondence, or size generally yields negligible gains or even substantial drops in other abilities at the same level. Specifically, \texttt{B+Dist.} increases \textbf{Geom.} abilities by +3.2, while decreasing \textbf{Motion}, \textbf{Rel.}, and \textbf{Local.} by -2.0, -5.8, and -4.6, respectively. However, it provides non-trivial gains in higher-level abilities, notably \textbf{Underst.} (+2.0) and \textbf{Goal Exec.} (+3.4). To give a further exploration on how this cability transfer happens and why, we provide a qualitative examples in Fig.~\ref{fig:ability_transfer}. After being fine-tuned on distance-QA data, \texttt{B+Dist.} can generalize to much more complex distance-related questions in in-the-wild scenarios, including those with novel coordinate prompts and multiple points queried simultaneously. This indicates that the model has learned an awareness of distance rather than overfitting to specific QA templates. Besides, and more intriguingly, the improved distance ability also shows clear cross-level transfer. It benefits higher-level tasks such as robot-arm manipulation, where MLLMs are required to guide the gripper to move, rotate, and open/close in 3D space. A better sense of metric space helps the model generate more reasonable control decisions in the real world.

\begin{tcolorbox}[
    colback=seedblue!5,
    colframe=seedblue!80,
    boxrule=0.5pt,
    arc=2pt,
    left=6pt,
    right=6pt,
    top=6pt,
    bottom=6pt,
    fonttitle=\bfseries,
    title=\textbf{Finding 2}
]
\textbf{Multi-ability Synergy:} 
The holistic integration across multiple fundamental abilities achieves synergistic gains far exceeding their individual effects.
\end{tcolorbox}
Tab.~\ref{tab:sft_exp} reveals an interesting phenomenon: individual SFT on any single ability—Distance, Size, or Correspondence—has limited impact on overall spatial performance, and can even slightly reduce it (e.g., \texttt{B+Dist.} \textbf{-0.5}, \texttt{B+Corr.} \textbf{+0.2}, \texttt{B+Size.} \textbf{-1.5} relative to the baseline). In contrast, combining all three abilities in a blended SFT (\texttt{B+Dist.+Size+Corr.}) yields an overall gain of \textbf{+1.1}, surpassing the performance of any individual ability and even exceeding the sum of their separate contributions. Remarkably, for abilities that suffered substantial drops under single-ability SFT—such as L1.Motion (best individual change \textbf{-2.0})—the compositional training produces a positive improvement of \textbf{+0.7}.

\subsection{Reinforcement Learning}
\begin{table*}[ht!]
    \centering
    \resizebox{\linewidth}{!}{
    \begin{tabular}{l|c|ccccc|cc|cc|cc}
    
    \hline 
    
    \rowcolor{white} 
     & & 
    \multicolumn{5}{c|}{\cellcolor{L1_Purple} L1 Perception} &  
    \multicolumn{2}{c|}{\cellcolor{L2_Orange} L2 Mental Mapping} &
    \multicolumn{2}{c|}{\cellcolor{L3_Green} L3 Mental Simulation} &
    \multicolumn{2}{c}{\cellcolor{L4_Blue} L4 Agentic Competence} \\
    
    \rowcolor{white} 
    \textbf{Methods} & \textbf{Avg.} & 
    \textbf{Geom.} & \textbf{Motion} & \textbf{Rel.} & \textbf{Local.} & \textbf{Orient.} &
    \textbf{Underst.} & \textbf{Memory} &
    \textbf{Caus. Reas.} & \textbf{Seq. Plan.} &
    \textbf{Goal Exec.} & \textbf{Open Expl.} \\

    \hline

    \rowcolor{navyblue!5}
    Qwen2.5-VL-7B & 27.5 & 17.8 & 22.6 & 23.9 & 20.6 & 21.6 & 34.7 & 15.2 & 28.4 & 39.8 & 24.5 & 31.1 \\ 
    Full RL@think & 28.2 & \goodres{19.0}{1.2} & \goodres{25.2}{2.7} & \neutralres{23.6}{-0.4} & \neutralres{20.3}{-0.3} & \neutralres{31.5}{-0.1} & \badres{30.6}{-4.1} & \badres{13.1}{-2.1} & \goodres{33.6}{5.2} & \badres{26.8}{-13.0} & \goodres{29.7}{5.2} & \goodres{40.0}{9.0} \\ 
    \hline
    \multicolumn{13}{l}{\textcolor{black}{\textit{After SFT}}} \\
    SFT Baseline & 27.3 & 28.6 & 24.6 & 20.6 & 26.3 & 36.0 & 22.2 & 22.6 & 28.2 & 32.8 & 23.3 & 35.9 \\ 
    L1 RL@think & \neutralres{26.6}{-0.7} & \neutralres{28.4}{0.9} & \goodres{30.0}{5.4} & \goodres{30.5}{9.9} & \badres{19.5}{-6.8} & \badres{34.5}{-1.4} & \goodres{24.9}{2.8} & \badres{18.5}{-4.1} & \badres{25.7}{-2.4} & \goodres{34.0}{1.1} & \neutralres{24.1}{0.8} & \badres{29.6}{-6.3} \\ 
    L2 RL@think & \neutralres{26.7}{-0.5} & \badres{24.4}{-4.1} & \badres{22.7}{-1.9} & \goodres{22.3}{1.6} & \badres{17.0}{-9.4} & \neutralres{35.0}{-0.9} & \goodres{26.6}{4.5} & \badres{16.1}{-6.5} & \neutralres{29.0}{0.8} & \badres{31.4}{-1.4} & \goodres{25.6}{2.3} & \badres{34.5}{-1.4} \\ 
    L3 RL@think & \neutralres{27.7}{0.4} & \badres{16.2}{-12.3} & \neutralres{24.0}{-0.6} & \goodres{24.1}{3.5} & \badres{21.4}{-4.9} & \goodres{38.5}{2.5} & \goodres{26.0}{3.9} & \neutralres{21.8}{-0.8} & \goodres{31.3}{3.1} & \goodres{34.7}{1.9}& \goodres{26.5}{3.2} & \goodres{38.4}{2.5} \\ 
    L4 RL@think & \goodres{28.5}{1.2} & \badres{23.8}{-4.8} & \neutralres{25.3}{0.7} & \goodres{22.1}{1.4} & \badres{23.5}{-2.8} & \badres{33.9}{-2.1} & \goodres{25.1}{3.0} & \badres{20.5}{-2.2} & \goodres{32.0}{3.8} & \goodres{34.2}{1.4}& \goodres{27.1}{3.9} & \goodres{38.8}{2.9} \\ 
    Full RL@think & \goodres{30.1}{2.9} & \goodres{29.7}{1.1} & \neutralres{24.7}{0.1} & \goodres{27.2}{6.5} & \badres{21.0}{-5.3} & \badres{34.8}{-1.2} & \goodres{27.4}{5.2} & \badres{16.7}{-5.9} & \goodres{33.6}{5.5} & \goodres{37.6}{4.8} & \goodres{25.4}{2.1} & \goodres{41.7}{5.8} \\ 
    Full RL@auto-think & \goodres{30.8}{3.6} & \goodres{31.9}{3.3} & \goodres{28.6}{4.0} & \goodres{22.0}{1.3} & \badres{23.1}{-3.2} & \neutralres{36.8}{0.8} & \goodres{28.0}{5.8} & \neutralres{22.6}{-0.1} & \goodres{33.5}{5.4} & \goodres{35.6}{2.8} & \neutralres{23.4}{0.1} & \goodres{44.1}{8.3} \\ 
    \hline
    \end{tabular}
    } 
    \caption{\textbf{RLVR Comparisons.}
    The table compares the baseline Qwen2.5-VL-7B with the version enhanced by RL on Goal-Exec. tasks. Changes are color-coded as \textcolor{papergreen}{notable gains}, \textcolor{gray}{neutral influence}, and \textcolor{paperred}{drops}.}
    \label{tab:rl_exp}
\end{table*}

\label{sec:RL}
Building on the findings from Supervised Fine-Tuning (SFT), we further investigate the potential of Reinforcement Learning with Verifiable Rewards (RLVR) to scale spatial abilities. Specifically, we employ Group Relative Policy Optimization (GRPO)~\cite{shao2024deepseekmath} to align the model’s policy with the hierarchical nature of the SpatialTree. Our experiments reveal that a uniform RL strategy is insufficient, conversely a hierarchy-aware reward mechanism unlocks significant performance gains.

\noindent\textbf{Limitations of Naive RL on Low-Level Skills.} We initially conducted experiments using standard GRPO on individual capabilities selected from each level of the SpatialTree. The results exposed a critical limitations of applying naive RL to spatial tasks, particularly regarding generalization in low-level skills. When optimizing exclusively for single low-level abilities, the model tends to overfit to the specific reward signal. This results in siloed improvements that fail to generalize to other foundational skills and, more critically, provide negligible or even detrimental transfer to high-level capabilities.

\noindent\textbf{Ineffectiveness of Uniform Data Mixing.} Attempts to mitigate this by combining datasets from multiple levels or mixing all available data for a unified RL stage yielded only marginal gains. The model struggled to balance the diverse requirements of the benchmark, suggesting that a "one-size-fits-all" reinforcement strategy cannot effectively span the spectrum from atomic perception to complex agentic planning.

\noindent\textbf{Hierarchy-Aware Reward Mechanism.} These limitations led us to hypothesize that different levels of spatial intelligence require distinct cognitive modes during training. While high-level reasoning benefits from extensive test-time computation, low-level perception is inherently intuitive and should function as a "fast" system. To test this, we introduced a Hierarchy-Aware Reward mechanism that adjusts the training objective based on the capability level:
\begin{itemize}
    \item \textbf{For Intuitive Perception:} For tasks such as depth estimation, object counting, and orientation, we removed rewards for ``thinking processes'' and introduced a length penalty. This implicitly discourages the model from over-reasoning on direct visual signals, forcing it to rely on direct visual-text alignment.
    \item \textbf{For Complex Reasoning:} For nested tasks like navigation planning and causal reasoning, we retained and amplified rewards for explicit reasoning steps, encouraging the model to utilize more tokens for intermediate computation.
\end{itemize}

This hierarchy-aware strategy proved highly effective. As shown in Table~\ref{tab:rl_exp}, the model trained with adaptive rewards significantly outperforms both the baseline and the naive GRPO variants across the entire SpatialTree-Bench. This finding strongly validates the structure of our taxonomy: Spatial Intelligence is not a flat collection of tasks, but a structured hierarchy where foundational perception requires direct alignment, while higher-order competence demands deliberate reasoning.

\noindent\textbf{Strict Evaluation Setup.} It is important to note two critical factors in our experimental design that ensure the robustness of our findings:
\begin{itemize}
    \item Data Decontamination: The robotic arm data samples used for GRPO training are strictly separated from the SpatialTree-Bench testing data. There is no overlap in specific scenes or object configurations between the training and evaluation sets.
    \item Task and Metric Discrepancy: The training objective is purely maximizing the reward on discrete MCQ selection. In contrast, the SpatialTree-Bench evaluation employs a diverse set of continuous and semantic metrics (e.g., Mean Squared Error for distance, angular error for orientation, and execution success rates for agentic tasks).
\end{itemize}

\noindent\textbf{Results and Observations.} Despite the significant domain gap and the difference in task formulation, the GRPO-tuned Qwen2.5-VL-7B demonstrates notable improvements across multiple levels of the SpatialTree hierarchy compared to its base counterpart. This suggests that the model is not merely memorizing dataset-specific patterns, but is effectively internalizing generalized spatial reasoning policies through the reinforcement learning process. These preliminary findings highlight the potential of RLVR as a scalable pathway for advancing spatial intelligence in MLLMs.

\vspace{-1em}
\section{Conclusion and Future works}
We present SpatialTree, the first capability-centric framework for Spatial Intelligence, organizing abilities into four hierarchical layers. This structure enables analysis of how spatial abilities emerge, compose, and transfer across levels. It also opens opportunities to efficiently scale up spatial intelligence in MLLMs, by strategically leveraging different types of data: identifying which abilities are most effective for pre-training, which can be directly applied in reinforcement learning with minimal additional reasoning data during post-training, and which are acquired through real-world interactions. We believe this could provide a promising path toward advancing spatial intelligence in MLLMs.

\clearpage

\bibliographystyle{plainnat}
\bibliography{main}

\clearpage

\beginappendix

\renewcommand\thefigure{\Alph{figure}} 
\renewcommand\thetable{\Alph{table}}   
\renewcommand\thesection{\Alph{section}}   
\setcounter{figure}{0} 
\setcounter{table}{0} 
\setcounter{section}{0} 

\label{sec:appendix}

\tableofcontents

\newpage

\section{Visualization of Data Sources}
\begin{figure}[t]
    \centering
    \includegraphics[width=0.98\linewidth]{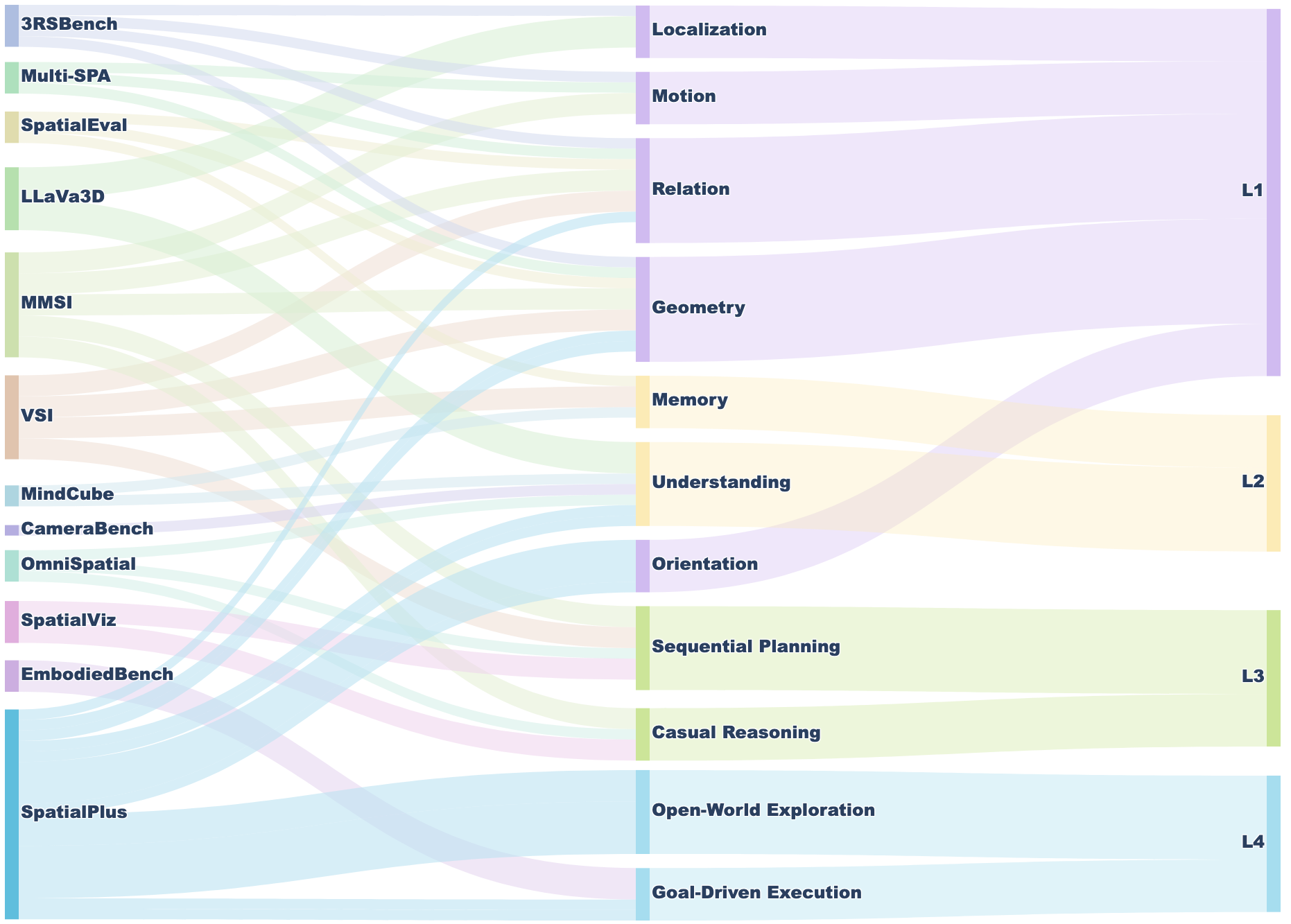}
   \caption{\textbf{Construction of SpatialTree-Bench.} We build our benchmark by reorganizing various existing datasets and mapping them to our capability tree, where \textbf{SpatialPlus}, a complementary dataset are introduced to ensure the capability coverage.}
    \label{fig:datasource}
\end{figure}

How different datasets contribute to our SpatialTree evaluation is shown in Fig.~\ref{fig:datasource}.

\section{Evaluation Metrics Details}

\paragraph{Multi-Option QAs.} 
For multi-option question answering, each model is evaluated on its ability to select the correct option from a predefined set. We measure accuracy by comparing the predicted choice against the ground-truth answer. This paradigm captures a model's understanding of spatial relations, object properties, and causal dynamics within a scene, corresponding to the low- and mid-level capabilities in the SpatialTree (L1–L3).

\paragraph{Numeric QAs.} 
Numeric QAs require models to predict continuous quantities such as distances, angles, or 3D coordinates. We evaluate performance using relative error metrics, for example: 
\[
\text{Relative Error} = \frac{|\hat{y}-y|}{|y|},
\] 
where $\hat{y}$ is the predicted value and $y$ is the ground truth. This metric ensures that predictions are scaled appropriately across different magnitudes and emphasizes precision in spatial reasoning.

\paragraph{GPT Judge.} 
For tasks that are open-ended or involve complex reasoning (e.g., trajectory description, action sequence explanation), we leverage a GPT-based judge to assess correctness. The judge evaluates whether the generated response satisfies the task requirements, optionally scoring partial correctness. This approach allows flexible evaluation beyond rigid numeric or multiple-choice formats, especially for mid- and high-level capabilities in L3–L4.

\paragraph{Agentic Evaluation.} 
To assess agentic competence, models are deployed in interactive simulated environments, such as those provided by EmbodiedBench~\citep{yang2025embodiedbench}. We evaluate navigation and manipulation tasks along multiple dimensions: success rate in completing the target goal, relative translation accuracy, and directional alignment. For each action step, a combined metric is computed using relative distance and cosine similarity of movement vectors, producing a normalized score in $[0,1]$. Aggregating scores over all steps yields a comprehensive measure of an agent’s ability to plan and execute actions in long-horizon, embodied tasks.

\section{SpatialPlus: Complementary Data Annotations for SpatialTree}

\begin{figure}[t]
    \centering
    \includegraphics[width=0.98\linewidth]{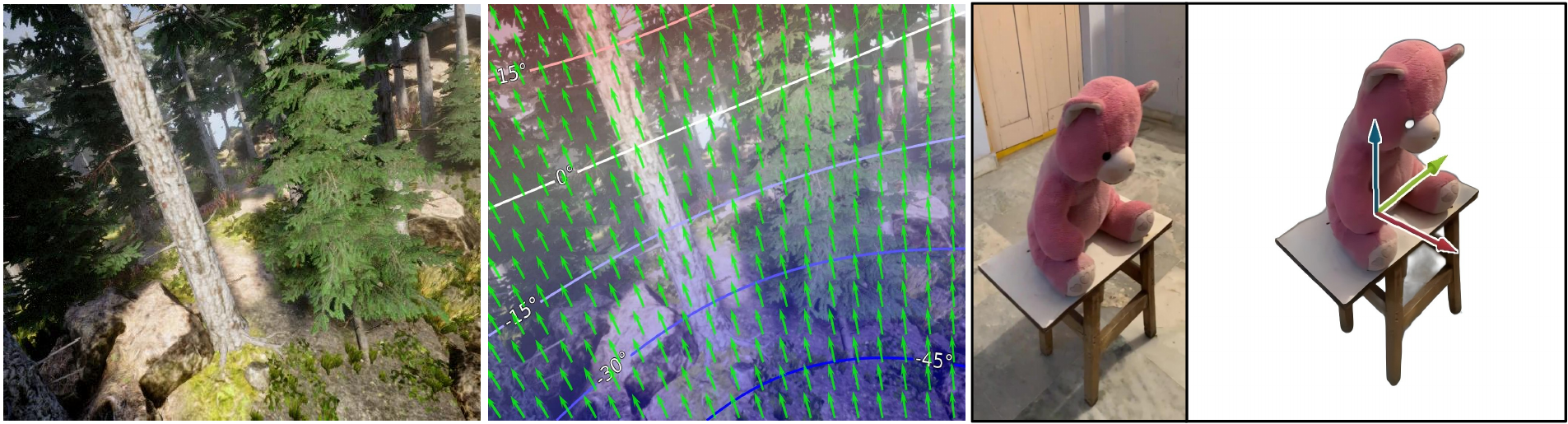}
    \caption{\textbf{Orientation Annotations.} The left side is the gravity field estimated from GeoCalib~\cite{veicht2024geocalib}, while the right side is from OrientAnything.}
    \label{fig:orientation}
\end{figure}

\subsection{Orientations (L1)}
\begin{figure}[H]
    \centering
    \includegraphics[width=0.99\linewidth]{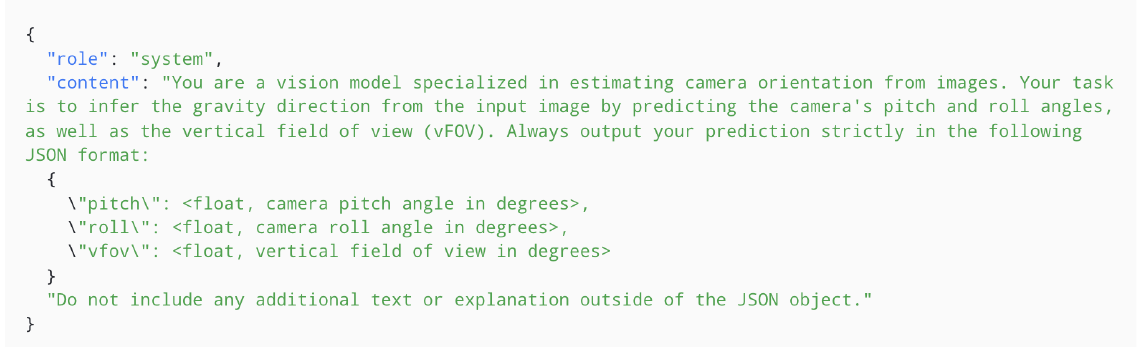}
    \caption{\textbf{Prompt template} for Orientation Estimation.}
    \label{fig:orientation_prompt}
\end{figure}

The Orientation capability, a fundamental yet under-explored area, involves estimating both gravity direction and 3D object orientation. To generate annotations, we leveraged Geocalib~\cite{veicht2024geocalib} for gravity vector estimation and OrientAnything~\cite{orient_anything} for object poses. We applied these tools to datasets suited for each task: for gravity, we annotated 500 images sampled from the diverse, drone-captured TartanAir~\cite{wang2020tartanair} dataset; for object orientation, we utilized the object-centric Co3dv2~\cite{reizenstein21co3d} dataset (Seen in Fig.~\ref{fig:orientation}). 
For gravity, the goal is to estimate the camera’s orientation relative to the gravity vector, typically represented by the \textit{pitch} and \textit{roll} angles. Formally, let the gravity vector in the world frame be:
\begin{equation}
\mathbf{g}_w = 
\begin{bmatrix}
0 \\ 0 \\ -1
\end{bmatrix},
\end{equation}
and let $\mathbf{R}_{cw} \in SO(3)$ denote the rotation from the world frame to the camera frame. The gravity direction in the camera frame is then:
\begin{equation}
\mathbf{g}_c = \mathbf{R}_{cw} \, \mathbf{g}_w.
\end{equation}
From $\mathbf{g}_c = [g_x, g_y, g_z]^\top$, the pitch and roll angles can be computed as:
\begin{align}
\text{pitch} &= \arctan2(-g_x, \sqrt{g_y^2 + g_z^2}), \\
\text{roll}  &= \arctan2(g_y, g_z).
\end{align}
Here, \textit{pitch} measures the forward--backward tilt of the camera, while \textit{roll} measures the sideways tilt. To evaluate an MLLM's proficiency in this task, we require the model to analyze the input image and return these same three parameters in a structured JSON format. An example of our prompt template is shown in Fig.~\ref{fig:orientation_prompt}.


For evaluation, we move beyond a simple absolute error metric and adopt a probabilistic approach that accounts for the inherent uncertainty of the ground-truth annotations provided by \textit{Geocalib}. For each predicted parameter (pitch, roll, and vFOV), \textit{Geocalib} also outputs an uncertainty value, which we interpret as the standard deviation ($\sigma_{gt}$). We then calculate a normalized similarity score ($S$) for each parameter using a Gaussian kernel, defined as:
\begin{equation}
\label{eq:uncertainty_score}
S(y_{\text{pred}}, y_{\text{gt}}, \sigma_{\text{gt}}) = \exp\left(-\frac{(y_{\text{pred}} - y_{\text{gt}})^2}{2\sigma_{\text{gt}}^2}\right)
\end{equation}
where $y_{\text{pred}}$ is the MLLM's prediction, $y_{\text{gt}}$ is the ground-truth value from \textit{Geocalib}, and $\sigma_{\text{gt}}$ is its associated uncertainty. This scoring function gracefully penalizes deviations from the ground truth: the score is 1 for a perfect match and decays towards 0 as the error increases. Crucially, a larger uncertainty $\sigma_{\text{gt}}$ in the ground truth leads to a slower decay, making the scoring more lenient when the ground truth itself is less certain. The final score for the task is the average of the individual scores for pitch, roll, and vFOV. For object orientation estimation, most of metrics are similar to gravity, and the evaluation are conducted on Azimuth, Polar and Rotation these three angles.

\subsection{Agentic Competence (L4)}
\label{appsubsec:agentic_com}

\begin{figure}[htbp]
    \centering
    \begin{subfigure}{0.365\linewidth}
        \centering
        \includegraphics[width=\linewidth]{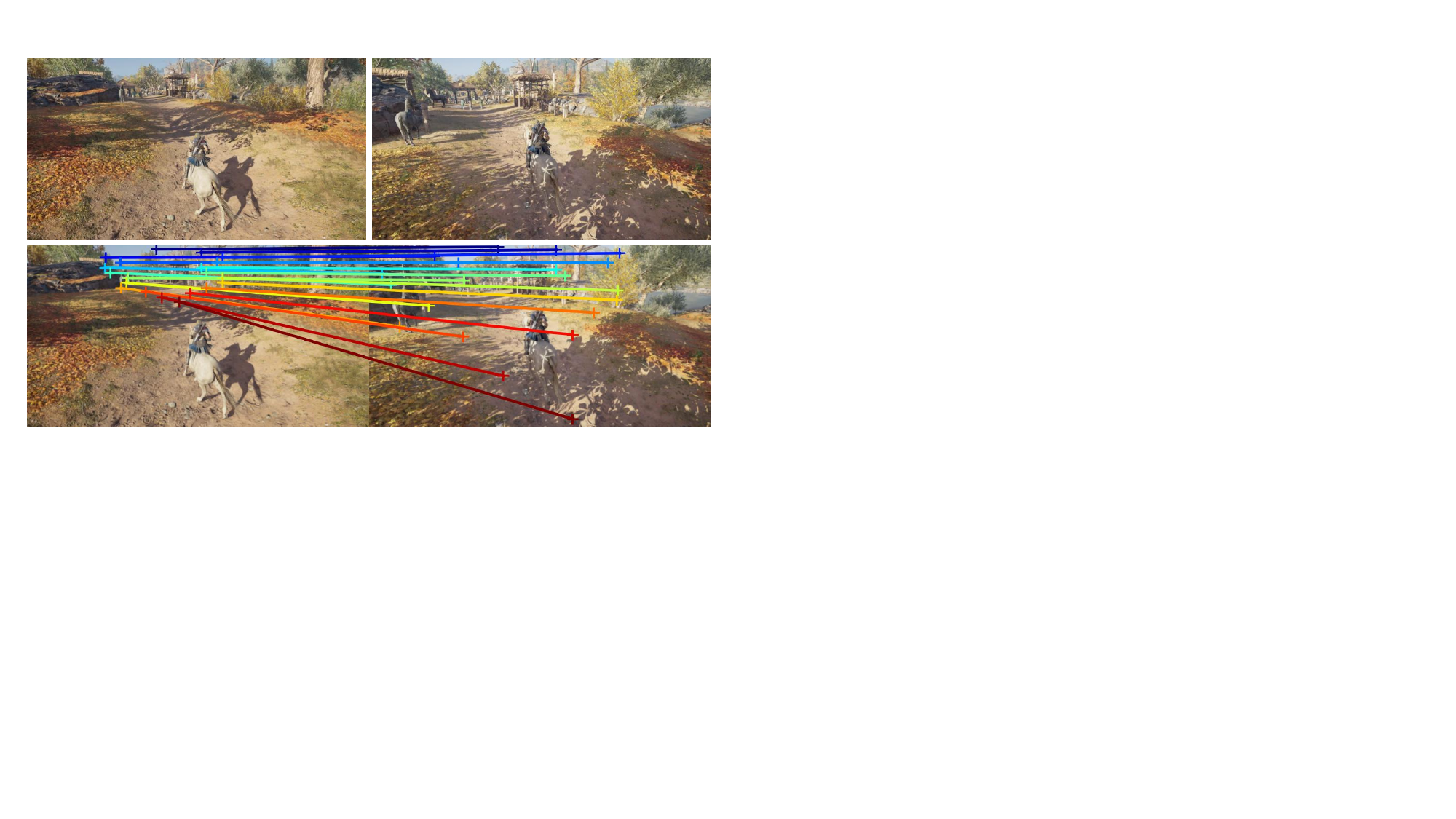}
        \caption{}
        \label{fig:sub_a}
    \end{subfigure}
    \hfill 
    \begin{subfigure}{0.6\linewidth}
        \centering
        \includegraphics[width=\linewidth]{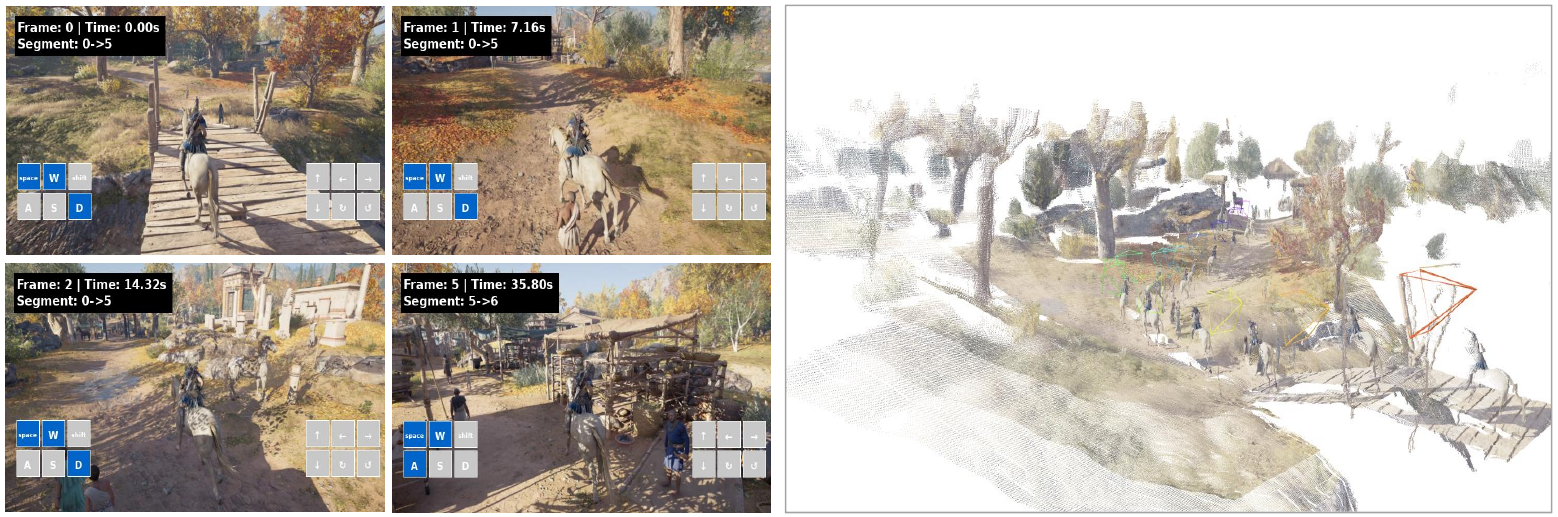}
        \caption{}
        \label{fig:sub_b}
    \end{subfigure}
    \caption{\textbf{Navigation Data Curation.} 
    (a) shows paired images used for evaluation, where MLLMs are expected to move from left to right. 
    (b) illustrates our curation process: reconstructing metric 3D models and camera trajectories, then converting them into actions.}
    \label{fig:navigation}
\end{figure}

\textbf{Spatial Action Mapping.} 
In the context of spatial agents, navigation and manipulation represent the most common forms of interaction within 3D environments. We address each with a distinct action space design. For navigation, we conceptualize agent movement as a series of camera motion controls (referring to recent video world models~\citep{genie3,yume}). To enable precise and intuitive control, we decompose complex camera movements into a set of fundamental motion primitives inspired by established cinematography techniques. This approach allows us to translate high-level language instructions (e.g., "move to the left," "look up") into a structured, low-level action space. The six fundamental primitives, their corresponding cinematic terms, degrees of freedom (DoF), and parameterization are detailed in Tab.~\ref{tab:spatial_action_mapping}.
\begin{table}[t!]
\centering
\caption{\textbf{Spatial Action Mapping.} This table defines a standardized interface that maps continuous 6-DoF motions and discrete control signals into action primitives with unified parameterization, enabling MLLMs to plan and execute embodied behaviors for agentic competence evaluation.}
\label{tab:spatial_action_mapping}
\vspace{-0.5em}
\small
\resizebox{\columnwidth}{!}{%
\begin{tabular}{@{}lll>{\centering\arraybackslash}m{3.0cm}>{\centering\arraybackslash}m{2.2cm}cc@{}} 
\toprule
\textbf{Primitive} & \textbf{Primitive Term} & \textbf{Category} & \textbf{Description} & \textbf{Action Mapping} & \textbf{Param.} & \textbf{Threshold} \\
\midrule
$P_{\text{truck}}$ & Truck & Translation & Move camera left/right (X-axis) & $A / D$ & $v_x$ & $\pm 0.01$ m/s \\
$P_{\text{dolly}}$ & Dolly & Translation & Move camera forward/backward (Z-axis) & $W / S$ & $v_z$ & $\pm 0.01$ m/s \\
$P_{\text{pedestal}}$ & Pedestal & Translation & Move camera up/down (Y-axis) & $Q / E$ & $v_y$ & $\pm 0.01$ m/s \\
$P_{\text{pan}}$ & Pan & Rotation & Turn camera left/right (yaw) & $\leftarrow / \rightarrow$ & $\omega_y$ & $\pm 0.5^\circ$/s \\
$P_{\text{tilt}}$ & Tilt & Rotation & Tilt camera up/down (pitch) & $\uparrow / \downarrow$ & $\omega_x$ & $\pm 0.5^\circ$/s \\
$P_{\text{roll}}$ & Roll & Rotation & Roll camera CW/CCW (roll) & $Z / X$ & $\omega_z$ & $\pm 0.5^\circ$/s \\
\midrule
$O_{\text{gripper}}$ & Gripper & Gripper Control & Open or close the gripper & $G / H$ & State $\in \{0,1\}$ & N/A \\
$O_{\text{push/pull}}$ & Push/Pull & Gesture & Push or pull object along forward axis & $P / L$ & Dir. $\in \{-1,+1\}$ & N/A \\
$O_{\text{grab}}$ & Grab & Gesture & Grab or release object & None & State $\in \{0,1\}$ & $G / H$ \\
\bottomrule
\end{tabular}}
\vspace{-5mm}
\end{table}

Formally, the camera trajectories are defined with a series of 
Camera-to-World (C2W) transformation matrices
$\mathcal{T}_{\text{motion}} = \{\mathbf{T}|_{0}^{i}, i=0,1,\dots,t\}$,
while the camera transformation at each moment is 
$\mathcal{T}_{i \rightarrow {i+1}} = \textbf{T}_{i+1}\textbf{T}_{i}^{-1}, 
\; i=0,1,\dots,t-1$. Then the continuous camera transformation can be decomposed into different components corresponding to each motion primitive, and discretized into the navigation action $A_{\text{nav}}$ using a speed threshold: 
\begin{align}
\label{eq:discrete_navigation_action}
\mathbf{A}_i^{\text{nav}} 
= \mathbf{T}_{i \rightarrow i+1} 
& = \{\Delta \mathbf{R}, \Delta \mathbf{t}\} \\
& \approx 
\left\{ \begin{array}{l}
t_i \cdot v_i, \; t_k \cdot\omega_k \mid i,k \in \{x,y,z\}, \\
t_i, t_k \in \mathbb{Z}_{\ge 0} 
\end{array} \right\} \nonumber
\end{align}
where $\Delta \mathbf{R} = (\Delta R_x, \Delta R_y, \Delta R_z)$ represents the rotation components obtained via Euler decomposition, $\Delta \mathbf{t} = (\Delta t_x, \Delta t_y, \Delta t_z)$ denotes the translation components along the $x$, $y$, and $z$ axes, and $t_i, t_k$ are discrete integers ranging from 0 up to the video frame rate (FPS). For manipulation, we focus on two representative scenarios to simplify the problem and enable controlled evaluation: human-hand manipulation and robotic gripper manipulation. For the gripper setting, we include gripper open/close actions along with wrist-level 6-DoF motion. For the human-hand setting, we define a small set of intuitive gesture primitives (i.e., push, pull, grab) seen in Tab.~\ref{tab:spatial_action_mapping} that capture essential interaction patterns. These manually defined mappings create a unified yet tractable action space for analyzing MLLMs' planning and manipulation competence.

Building on the proposed spatial action mapping, we curate annotated data from diverse sources, including human-hand manipulation videos, navigation video games, robotic arm manipulation datasets, and simulation environments. This unified dataset enables us to evaluate whether MLLMs can accurately plan and execute actions in the defined metric action space. Further implementation details are provided in Sec.~\ref{sec:si-engine} and in the experimental section.

\textbf{Goal-driven Navigation.}
We leverage our SpatialEngine to get the action annotations as shown in Fig.~\ref{fig:navigation}. We first extract the metric pose trajectories from the games videos, and convert them into discrete actions with our spatial action mapping, and then we randomly sample several image pairs from the video with the correspondence checking. For evaluation, the goal is a image, and the MLLMs are supposed to control the character to move to the target positions. We use the prompt template as below:

\begin{figure}[H]
    \centering
    \includegraphics[width=0.99\linewidth]{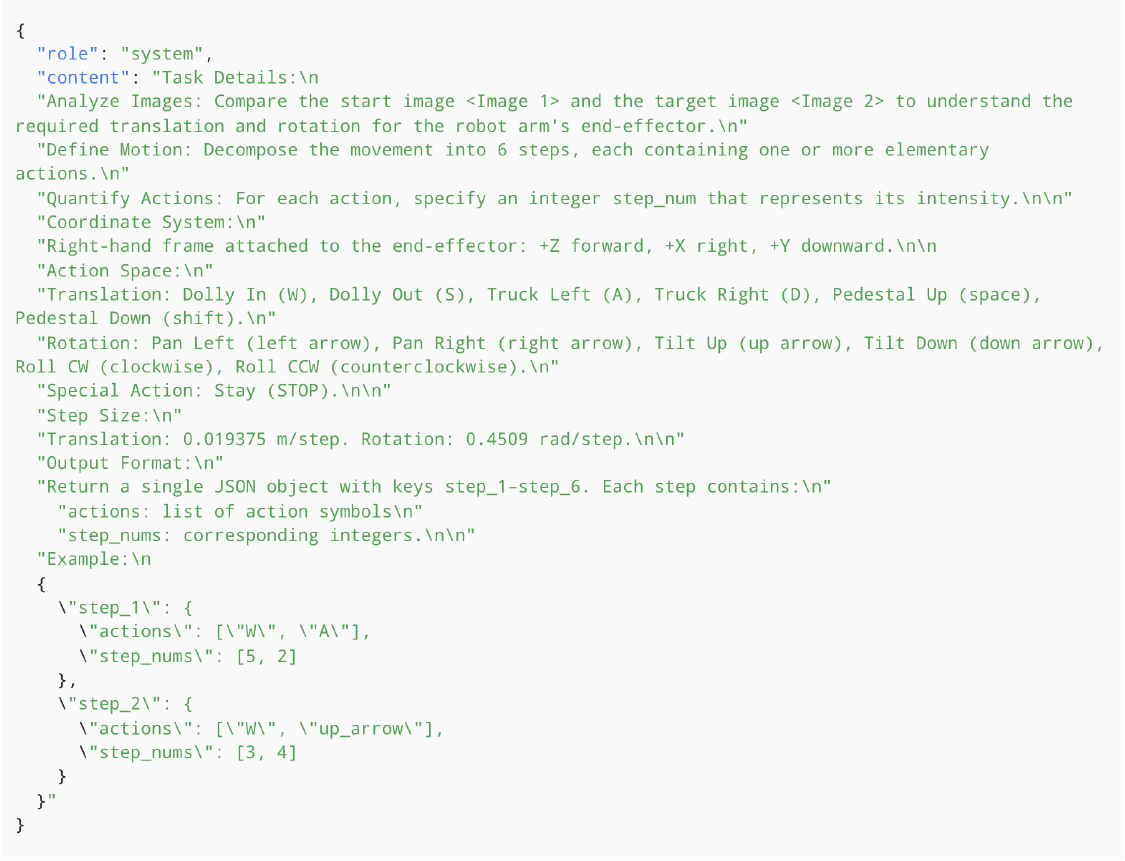}
    \caption{\textbf{Prompt of navigation.}}
    \label{lst:navigation_p}
\end{figure}

In this prompt, translation and rotation steps are computed from the actual movement, while capping the number of steps at 10 to prevent overly long action sequences. To evaluate MLLMs, we compute a normalized metric in the range $[0, 1]$ by combining \textbf{relative distance} and \textbf{directional accuracy}. Specifically, for each step, let $\Delta \mathbf{p}_{\text{pred}}$ and $\Delta \mathbf{p}_{\text{gt}}$ denote the predicted and ground-truth translation vectors, respectively. 

The \textbf{relative distance score} is defined as:
\[
s_d = \max\Big(0, 1 - \frac{\|\Delta \mathbf{p}_{\text{pred}} - \Delta \mathbf{p}_{\text{gt}}\|}{\|\Delta \mathbf{p}_{\text{gt}}\|}\Big),
\]

and the \textbf{directional score} is computed by the cosine similarity:
\[
s_\theta = \frac{\Delta \mathbf{p}_{\text{pred}} \cdot \Delta \mathbf{p}_{\text{gt}}}{\|\Delta \mathbf{p}_{\text{pred}}\| \, \|\Delta \mathbf{p}_{\text{gt}}\|}.
\]

The final step-wise accuracy is then: $s_{\text{step}} = s_d \cdot \max(0, s_\theta)$

which ensures a value in $[0,1]$, where 1 indicates perfect alignment in both distance and direction. Aggregating $s_{\text{step}}$ across all steps provides a comprehensive measure of the model's precision in executing end-effector motions.

\textbf{Goal-driven Manipulation}
For the \textbf{Goal-Driven Manipulation} capability, we utilize action annotations from the \texttt{Droid}~\cite{khazatsky2024droid} and \texttt{EgoDex}~\cite{egodex} datasets. This task requires the MLLM to generate a sequence of precise actions to move a robot end-effector or a human hand from a starting state to a target state, both specified by images. The action space for \texttt{Droid} encompasses 7-DoF control: 6-DoF for the end-effector's pose (translation and rotation) and a binary state for the gripper (open/close). A similar action space is adapted for \texttt{EgoDex}, controlling wrist pose and finger grip. The MLLM is prompted to generate a sequence of continuous action vectors, as shown in the template below:

\begin{figure}[H]
    \centering
    \includegraphics[width=0.99\linewidth]{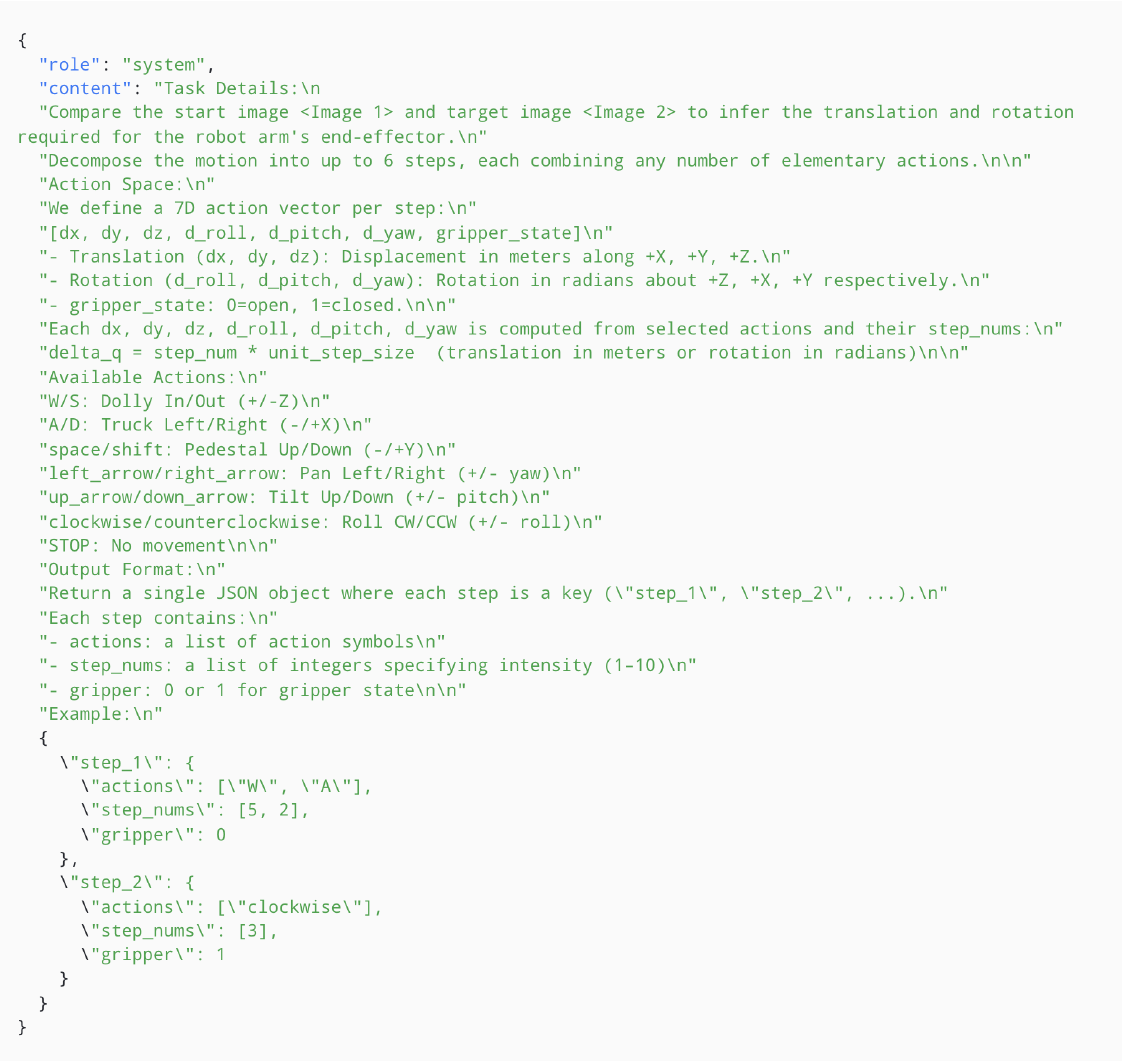}
    \caption{\textbf{Prompt for Goal-Driven Manipulation with 7D Action Representation.}}
    \label{lst:manipulation_p}
\end{figure}

To evaluate the MLLM's performance, we assess the accuracy of the predicted action sequence against the ground truth. For the translational component of the motion, we reuse the step-wise accuracy metric $s_{\text{step}}$ from the navigation task, which combines relative distance and directional scores. For the rotational component, we compute a normalized score based on the angular difference between the predicted orientation and the ground truth. Let $R_{\text{pred}}$ and $R_{\text{gt}}$ be the predicted and ground-truth rotation matrices for a step. The rotational error angle $\theta_{\text{err}}$ is calculated from the error rotation matrix $R_{\text{err}} = R_{\text{pred}} R_{\text{gt}}^T$:
\[
\theta_{\text{err}} = \arccos\left(\frac{\text{Tr}(R_{\text{err}}) - 1}{2}\right).
\]
The \textbf{rotation score} $s_{\text{rot}}$ is then defined as:
\[
s_{\text{rot}} = \max\Big(0, 1 - \frac{\theta_{\text{err}}}{\pi}\Big),
\]
which normalizes the error to a $[0, 1]$ range, where 1 indicates a perfect rotational match. Finally, the \textbf{gripper score} $s_{\text{gripper}}$ is a binary accuracy (1 if the predicted state matches the ground truth, 0 otherwise). The final score for each step is a weighted combination of these three metrics, providing a holistic evaluation of the model's ability to perform precise, multi-faceted manipulation tasks.



\section{Ability Transfer via Prompting}
\label{sec:atomic_prompting}

\begin{figure}[htbp]
\centering
\includegraphics[width=1\linewidth]{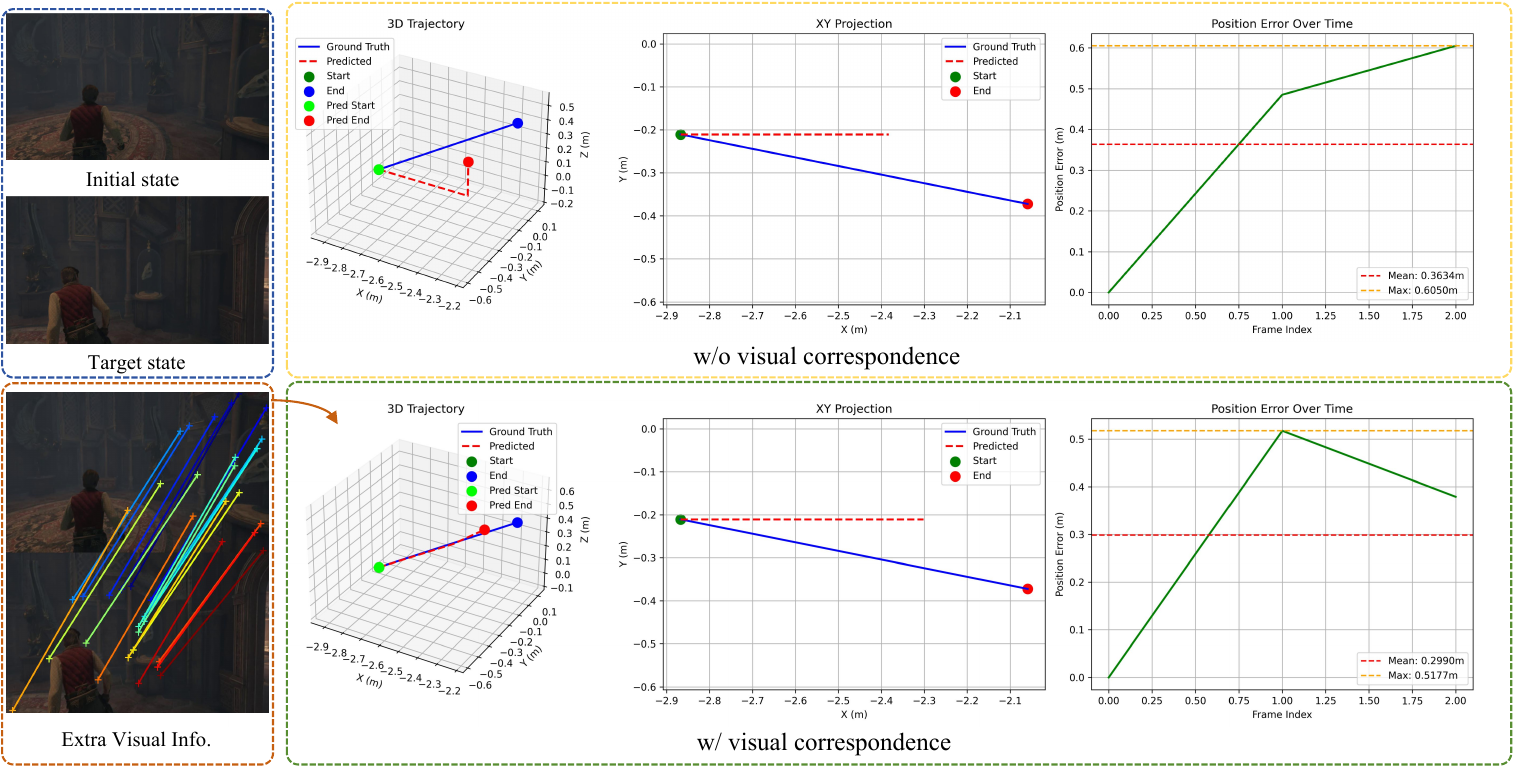}
\vspace{-1em}
\caption{\textbf{Correspondence Prompting for Navigation.} The correspondence prompt guides Gemini2.5-pro to navigate and move  more accurately within 3D environments.}%
\vspace{-1em}
\label{fig:prompt}
\end{figure}

In addition to SFT, we investigate cross-level ability influence through explicit prompting. Specifically, we consider a representative task pair: low-level abilities (L1.Corr, L1.Dist, L1.Size) and a high-level task (L4.Imaged Goal Navigation). Intuitively, correspondence is a necessary component for navigation. Using Gemini2.5-pro, we provide models with explicit prompts derived from matching visualizations, depth, and size context. As shown in Fig.~\ref{fig:prompt}, correspondence guidance improves target direction recognition, increasing accuracy by \textbf{7.1\%}, while distance and size prompting yield gains of \textbf{5.5\%} and \textbf{2.1\%}, respectively. These results suggest that grounding MLLM reasoning with explicit low-level visual information can substantially enhance performance on complex spatial navigation tasks.
%

\section{Benchmark Metric Aggregation}

To derive a single, comprehensive score for a model's spatial intelligence, we employ a hierarchical aggregation methodology. This approach is designed to reflect the complex, multi-layered nature of spatial cognition, rather than treating all abilities as equally important. The design is principally guided by established theories in cognitive psychology, which posit that spatial intelligence is constructed hierarchically, with fundamental perceptual skills forming the bedrock for more abstract reasoning and planning.

Our aggregation framework is built upon the \textit{SpatialTree} structure. The assignment of weights within this tree is determined by a synthesis of theoretical principles and empirical, data-driven insights:

\begin{figure}[htbp]
\centering
\includegraphics[width=0.98\linewidth]{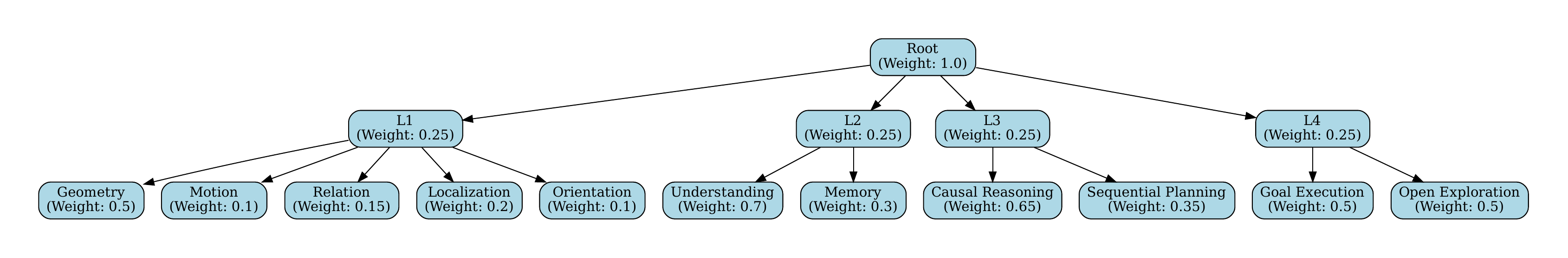}
\caption{An illustration of the hierarchical weighting scheme for metric aggregation with in the \textit{SpatialTree}. Each node represents a capability layer, with the assigned weight used for the bottom-up calculation of the final score. The weighting prioritizes foundational perceptual abilities (L1) as they are prerequisites for higher-level cognitive tasks.}%
\label{fig:weighted_tree}
\end{figure}

\textbf{Cognitive Hierarchy.} In line with cognitive science literature, our weighting scheme prioritizes foundational capabilities, as shown in Fig.~\ref{fig:weighted_tree}. The L1 layer, which represents low-level spatial perception, is assigned the largest weight, as these skills are prerequisites for almost all higher-level spatial tasks found in L2 (Mental Mapping) and L3 (Mental Simulation).

\textbf{Empirical Dependency from Correlation Analysis.} The theoretical hierarchy is further refined and validated by our empirical findings from the Pearson correlation heatmap (Fig.~\ref{fig:wrap_example}). The heatmap allows us to identify \textit{atomic abilities} that exhibit strong, widespread correlations with a multitude of other skills. These influential abilities are considered more fundamental to the overall spatial intelligence network and are consequently assigned higher weights within their respective sub-trees. This ensures our metric is not just theoretically sound, but also reflects the actual dependencies observed in model performance.

The final score is calculated via a bottom-up, weighted summation. The performance score for any parent node in the tree is the weighted sum of the scores of its immediate children. This process is recursively applied until the root node is reached, yielding a single, principled score that holistically quantifies the spatial intelligence of a given MLLM.

\section{More Visualizations for QAs in SpatialTree Bench}
\label{sec:vis_qa}
We declare that Large Language Models (LLMs) were used in a limited capacity during the preparation of this manuscript. Specifically, LLMs were employed for grammar checking, word choice refinement, and typo correction. All core technical contributions, experimental design, analysis, and conclusions are entirely our own. The use of LLMs did not influence the scientific methodology, result interpretation, or theoretical contributions of this research.

\begin{table*}[t]
\begin{minipage}{1.0\linewidth}
\centering
\tablestyle{4.0pt}{1.0}
\begin{tabular}{p{0.11\linewidth} p{0.85\linewidth}}
\toprule
 \multicolumn{2}{l}{\bf Geometry}  \\
\midrule
&  \includegraphics[width=0.27\textwidth]{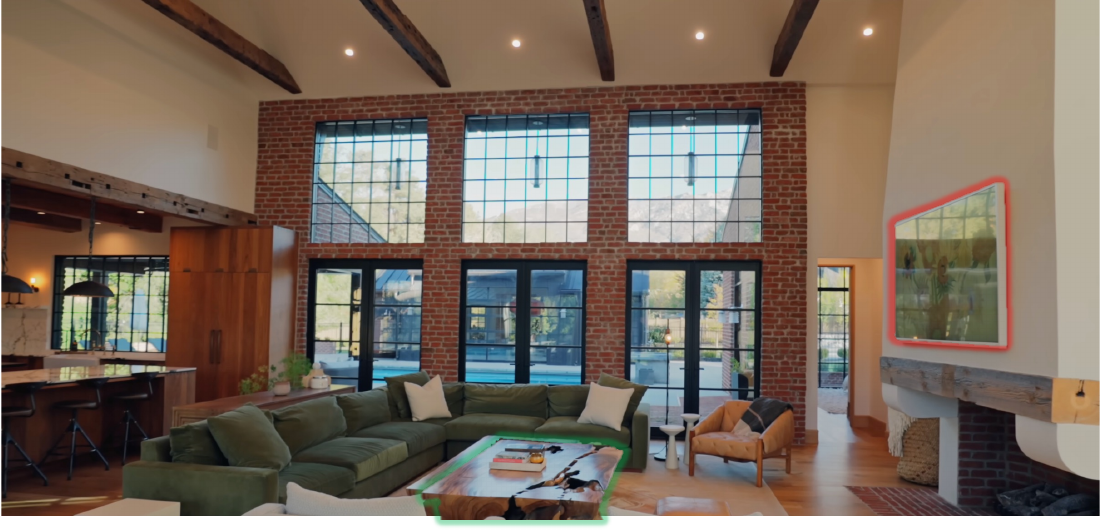}
\\
\textbf{Prompt} & Which is wider, the width of the painting on the wall or the width of the wooden table next to the sofa?
\\\cmidrule{1-2}
\textbf{Answer} & The wooden table
\\\midrule
&  \includegraphics[width=0.27\textwidth]{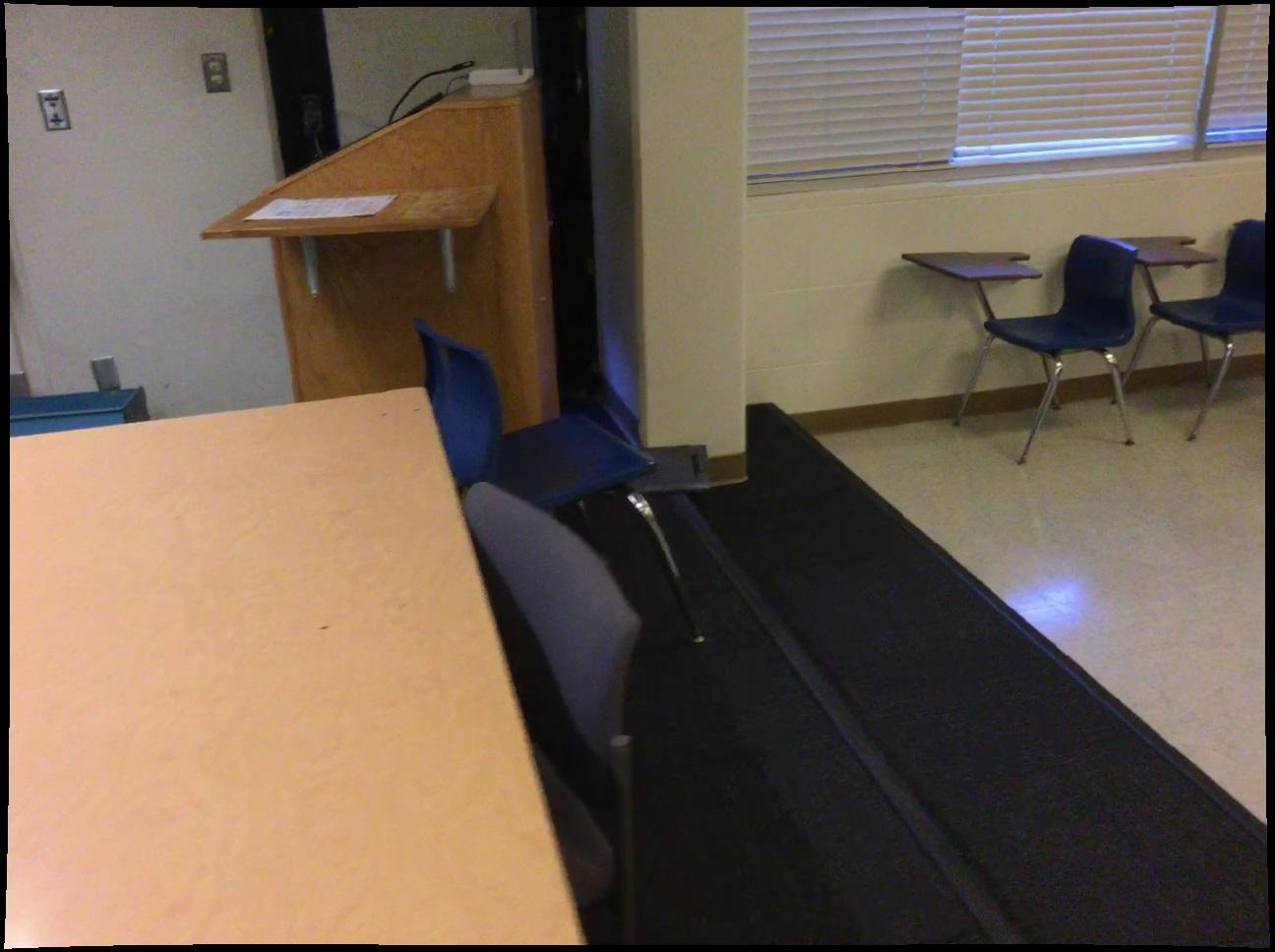}
\\
\textbf{Prompt} & The coordinates [ x , y ] are normalized to 0-1 and scaled by 1000, with [ 0 , 0 ] at the top-left. The x-axis represents the width, and the y-axis represents the height.
What is the depth at the coordinates [ 389 , 180 ] in the image (in mm)?\\\cmidrule{1-2}
\textbf{Answer} & 
\begin{verbatim}
	1915
\end{verbatim}
\\
\midrule
&  \includegraphics[width=0.27\textwidth]{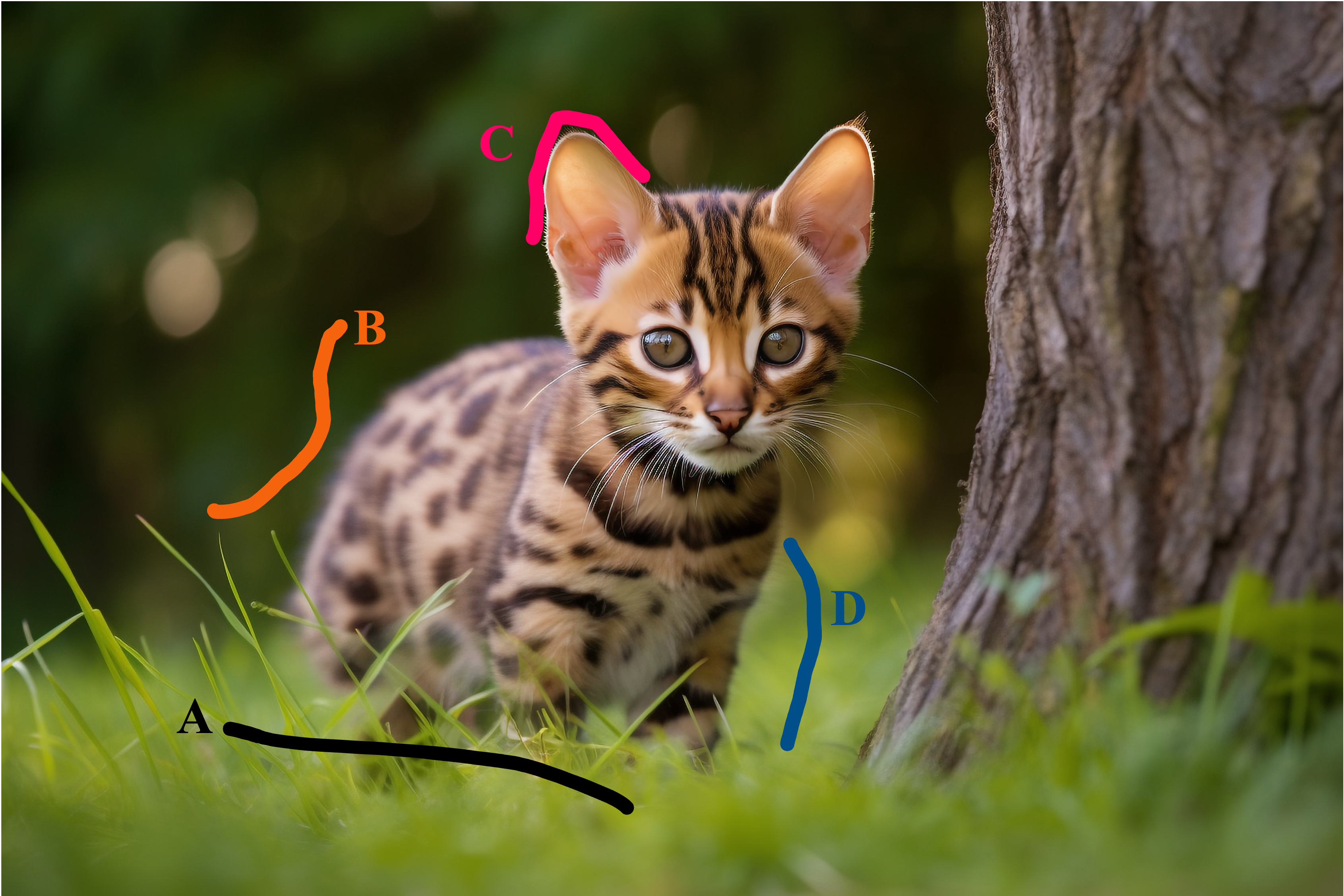}
\\
\textbf{Prompt} & Which line is closer to the edge?
\\\cmidrule{1-2}
\textbf{Answer} & C
\\
\bottomrule
\end{tabular}

\captionof{table}{Examples of \textit{L1.Geometry}.}
\label{tab:example_3dod_data}
\end{minipage}
\end{table*}

\begin{table*}[t]
\begin{minipage}{1.0\linewidth}
\centering
\tablestyle{4.0pt}{1.0}

\begin{tabular}{p{0.11\linewidth} p{0.85\linewidth}}
\toprule
 \multicolumn{2}{l}{\bf Relation}  \\
\midrule
&  \includegraphics[width=0.7\textwidth]{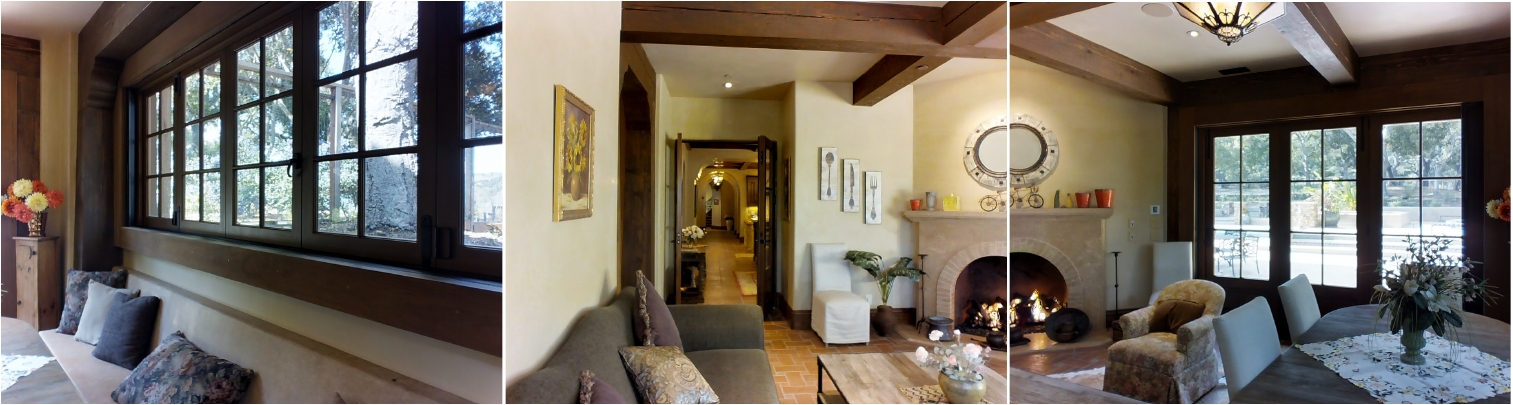}
\\
\textbf{Prompt} & When you were taking the photo in Image 1, where was the exit of the room relative to you?
\\\cmidrule{1-2}
\textbf{Answer} & Rear left
\\\midrule
&  \includegraphics[width=0.7\textwidth]{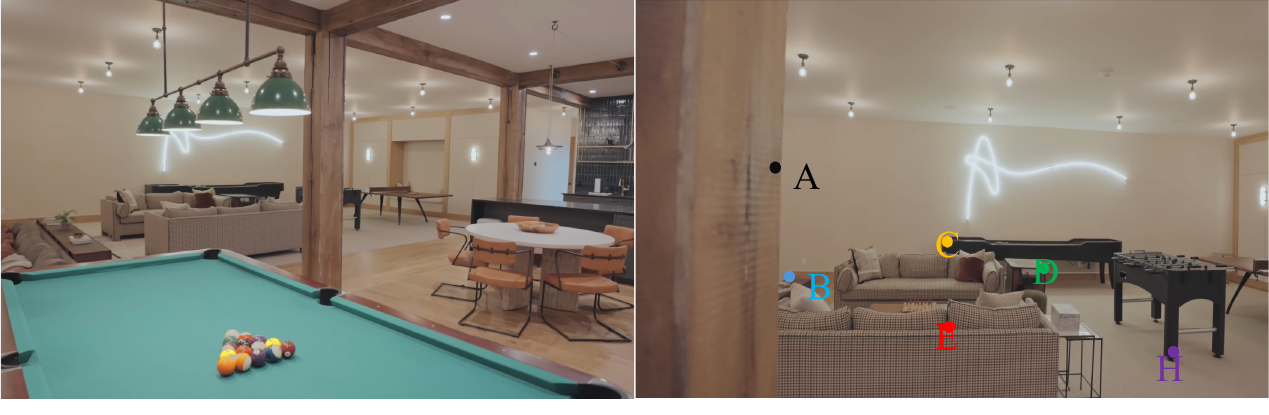}
\\
\textbf{Prompt} & Track [ 150,  470 ] from Image 1 to its correspondence in Image 2.
\\\cmidrule{1-2}
\textbf{Answer} & 
\begin{verbatim}
	C
\end{verbatim}
\\
\bottomrule
\end{tabular}

\captionof{table}{Examples of \textit{L1.Relation}.}
\label{tab:example_3dod_data}
\end{minipage}
\end{table*}

\begin{table*}[t]
\begin{minipage}{1.0\linewidth}
\centering
\tablestyle{4.0pt}{1.0}

\begin{tabular}{p{0.11\linewidth} p{0.85\linewidth}}
\toprule
 \multicolumn{2}{l}{\bf Orientation}  \\
\midrule
&  \includegraphics[width=0.27\textwidth]{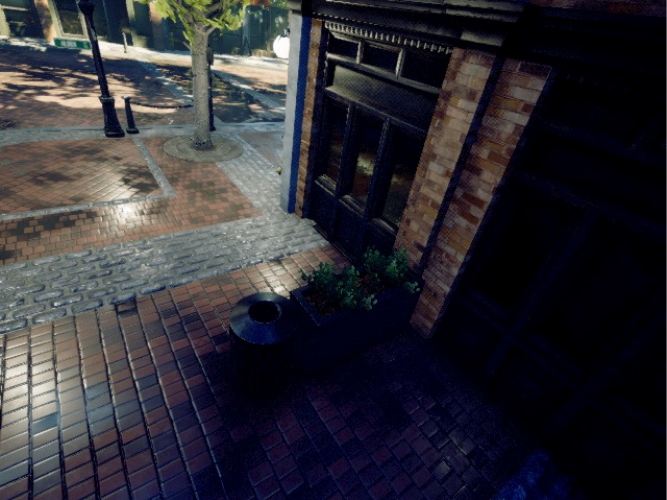}
\\
\textbf{Prompt} & Analyze the image to determine the vertical field of view (vfov) and calculate the camera'against the vertical axis (gras roll and pitch angles vity).
\\\cmidrule{1-2}
\textbf{Answer} & 
\begin{verbatim}[
"roll_unc": 0.61629718542099, 
"pitch_unc": 1.862020492553711, 
"vfov_unc": 20.077800750732422
]\end{verbatim}
\\\midrule
&  \includegraphics[width=0.27\textwidth]{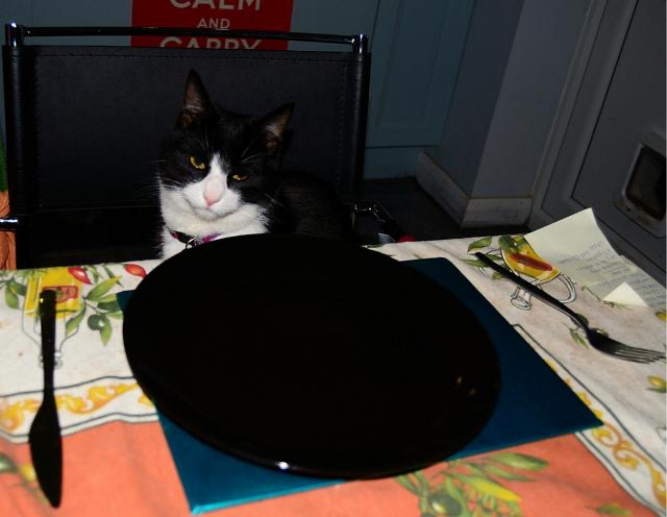}
\\
\textbf{Prompt} & If I stand at the cat's position facing where it is facing, is the knife in front of me or behind me?
\\\cmidrule{1-2}
\textbf{Answer} & In front of
\\
\bottomrule
\end{tabular}

\captionof{table}{Examples of \textit{L1.Orientation}.}
\label{tab:example_3dod_data}
\end{minipage}
\end{table*}

\begin{table*}[t]
\begin{minipage}{1.0\linewidth}
\centering
\tablestyle{4.0pt}{1.0}

\begin{tabular}{p{0.11\linewidth} p{0.85\linewidth}}
\toprule
 \multicolumn{2}{l}{\bf Motion}  \\
\midrule
&  \includegraphics[width=0.7\textwidth]{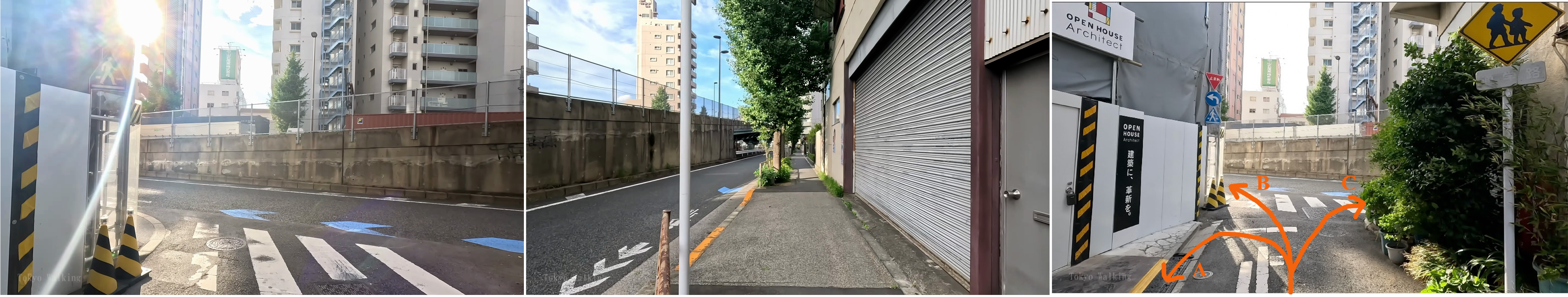}
\\
\textbf{Prompt} & As shown in the video, which direction is the view moving towards ?
\\\cmidrule{1-2}
\textbf{Answer} & C
\\\midrule
&  \includegraphics[width=0.7\textwidth]{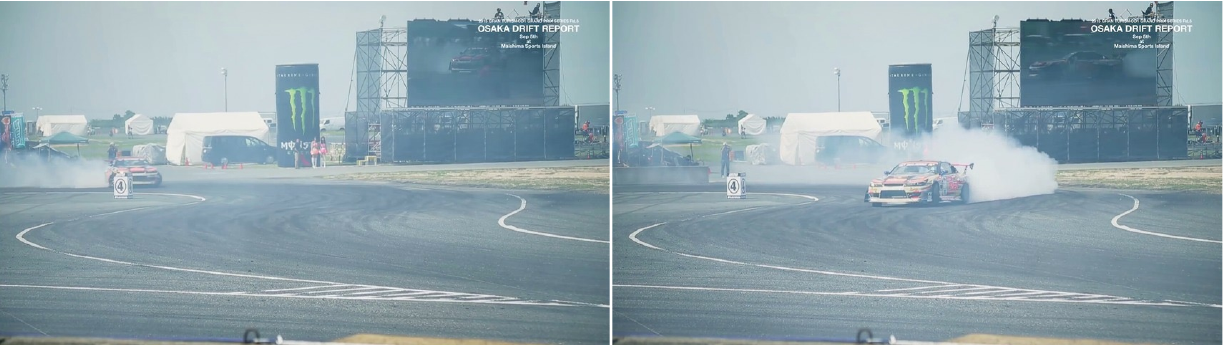}
\\
\textbf{Prompt} & From view 1 to view2, in which direction is the race car moving?
\\\cmidrule{1-2}
\textbf{Answer} & First to the front left, then to the front right.
\begin{verbatim}
	C
\end{verbatim}
\\
\bottomrule
\end{tabular}

\captionof{table}{Examples of \textit{L1.Motion}.}
\label{tab:example_3dod_data}
\end{minipage}
\end{table*}

\begin{table*}[t]
\begin{minipage}{1.0\linewidth}
\centering
\tablestyle{4.0pt}{1.0}

\begin{tabular}{p{0.11\linewidth} p{0.85\linewidth}}
\toprule
 \multicolumn{2}{l}{\bf Localization}  \\
\\\midrule
&  \includegraphics[width=0.27\textwidth]{figures/case/local-grounding.pdf}
\\
\textbf{Prompt} & Please ground mirror in this image. The 3D bounding box is defined in the format:
\begin{verbatim}
"bbox_3d": [u_center, v_center, z_center, x_size, y_size, z_size, roll, pitch, yaw]
\end{verbatim}

\\\cmidrule{1-2}
\textbf{Answer} & 
\begin{verbatim}[
	"bbox_3d": [901, 558, 4.87, 0.541, 1.698, 0.243, 179.279, -29.539, 176.489]
]\end{verbatim}
\\
\bottomrule
\end{tabular}

\captionof{table}{Examples of \textit{L1.Localization}.}
\label{tab:example_3dod_data}
\end{minipage}
\end{table*}

\begin{table*}[t]
\begin{minipage}{1.0\linewidth}
\centering
\tablestyle{4.0pt}{1.0}

\begin{tabular}{p{0.11\linewidth} p{0.8\linewidth}}
\toprule
 \multicolumn{2}{l}{\bf Understanding}  \\
\midrule
&  \includegraphics[width=0.27\textwidth]{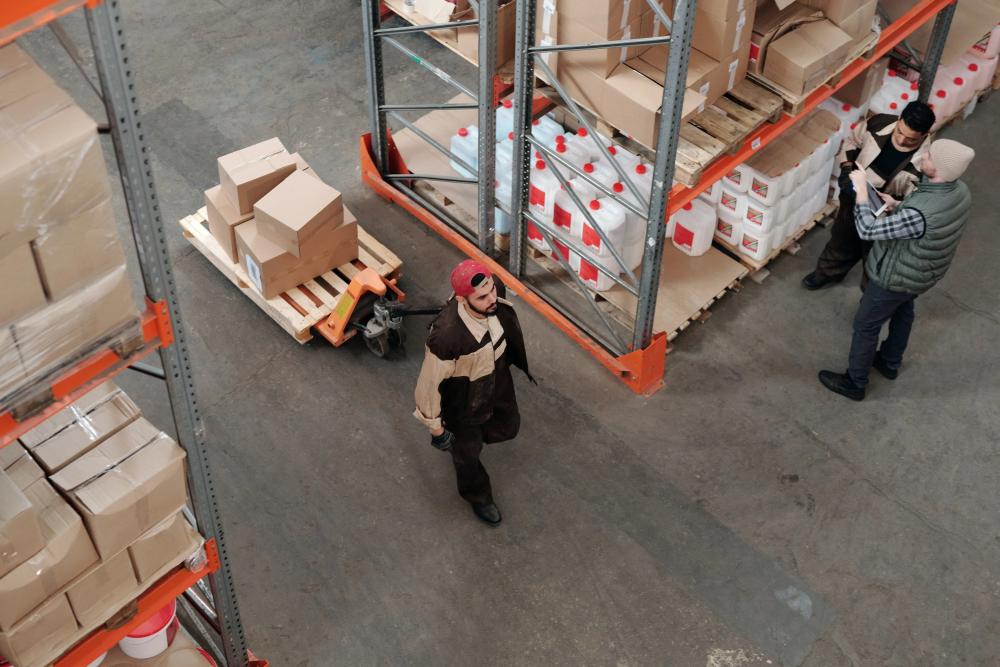}
\\
\textbf{Prompt} &From the man wearing the red hat, how many people are on his left?
\begin{verbatim}[
A. zero,
B. one,
C. three,
D. two
]\end{verbatim}
\\\cmidrule{1-2}
\textbf{Answer} & D
\\\midrule
&  \includegraphics[width=0.27\textwidth]{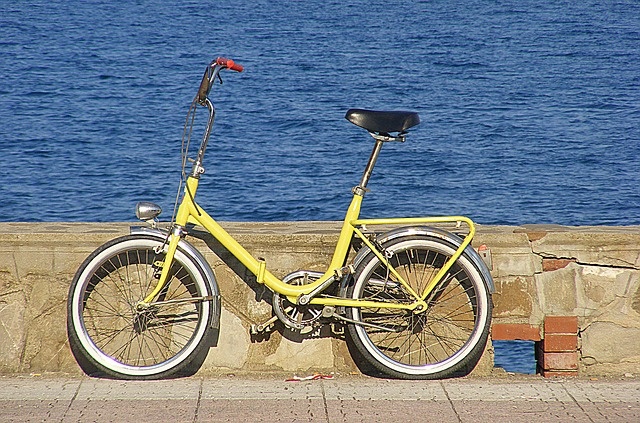}
\\
\textbf{Prompt} & To ride this bicycle along the seawall, where would you place your hands to steer, where would you sit, and where would you put your feet to pedal?
\\\cmidrule{1-2}
\textbf{Answer} &
\begin{verbatim}
	 [744, 439, 783, 489]
\end{verbatim}
\\
\bottomrule
\end{tabular}

\captionof{table}{Examples of \textit{L2.Understanding}.}
\label{tab:example_3dod_data}
\end{minipage}
\end{table*}

\begin{table*}[t]
\begin{minipage}{1.0\linewidth}
\centering
\tablestyle{4.0pt}{1.0}

\begin{tabular}{p{0.11\linewidth} p{0.8\linewidth}}
\toprule
 \multicolumn{2}{l}{\bf Memory}  \\
\midrule
&  \includegraphics[width=0.8\textwidth]{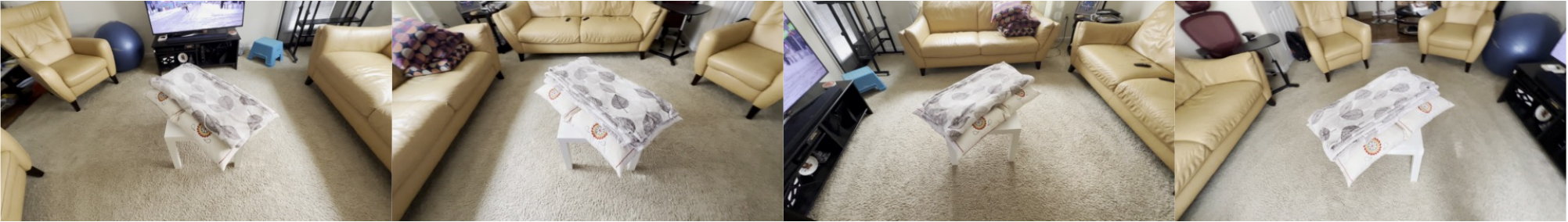}
\\
\textbf{Prompt} &From the man wearing the red hat, how many people are on his left?
\begin{verbatim}[
        A. Leather loveseat with three seat cushions,
        B. TV,
        C. Two single sofas,
        D. Leather loveseat
]\end{verbatim}
\\\cmidrule{1-2}
\textbf{Answer} & B
\\\midrule
&  \includegraphics[width=0.8\textwidth]{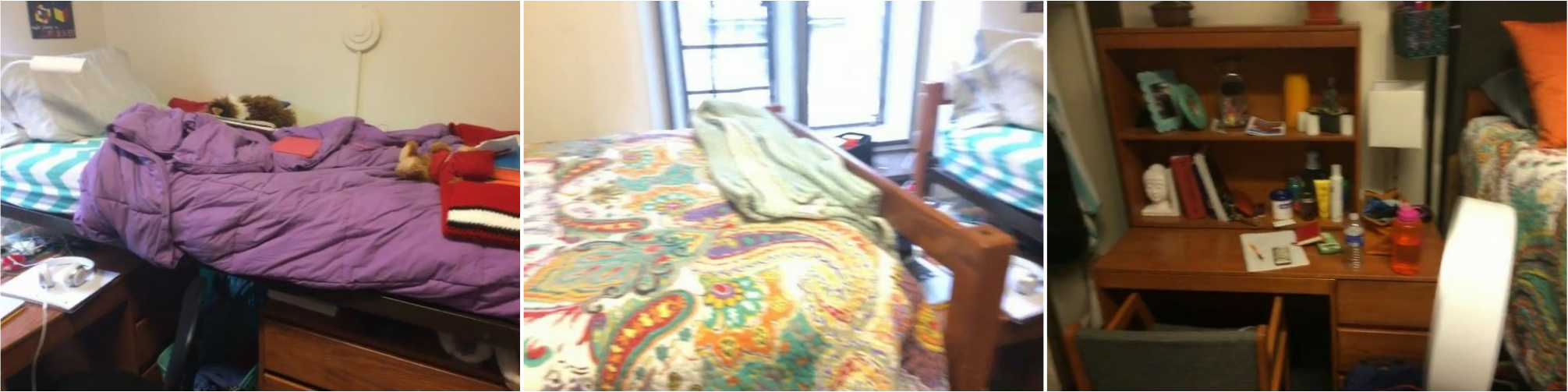}
\\
\textbf{Prompt} & How many bed(s) are in this room?
\\\cmidrule{1-2}
\textbf{Answer} & 2
\\
\bottomrule
\end{tabular}

\captionof{table}{Examples of \textit{L2.Memory}.}
\label{tab:example_3dod_data}
\end{minipage}
\end{table*}

\begin{table*}[t]
\begin{minipage}{1.0\linewidth}
\centering
\tablestyle{4.0pt}{1.0}

\begin{tabular}{p{0.11\linewidth} p{0.8\linewidth}}
\toprule
 \multicolumn{2}{l}{\bf Causal Reasoning}  \\
\midrule
&  \includegraphics[width=0.27\textwidth]{figures/case/L3_caus_dyn.jpg}
\\
\textbf{Prompt} &If the dog on the right reaches the camera in 5 s, what is its speed?
\begin{verbatim}[
        A. 14.7m/s
        B. 1.9m/s
        C. 21.7m/s
        D. 11.9m/s
]\end{verbatim}
\\\cmidrule{1-2}
\textbf{Answer} & B
\\\midrule
&  \includegraphics[width=0.27\textwidth]{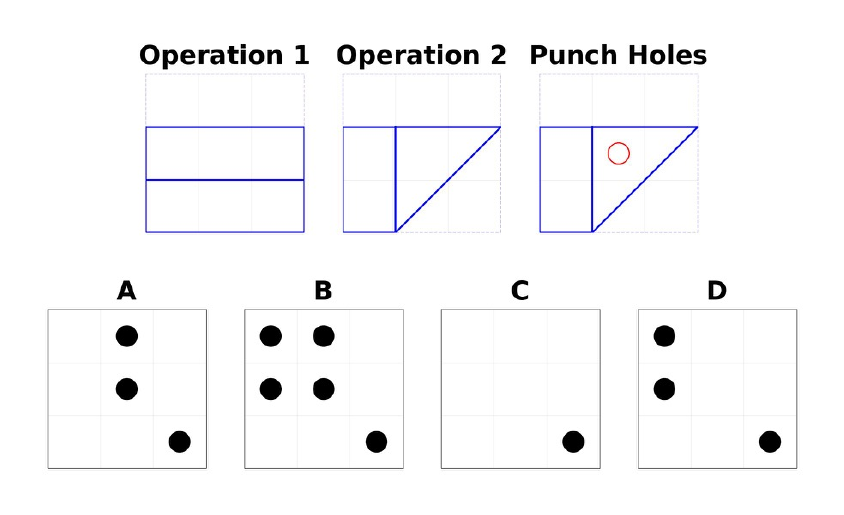}
\\
\textbf{Prompt} & A 3x3 grid paper undergoes two folds as shown. A hole is punched in the folded state. Which option (A, B, C, or D) shows the unfolded paper?
\\\cmidrule{1-2}
\textbf{Answer} & A
\\
\midrule
&  \includegraphics[width=0.27\textwidth]{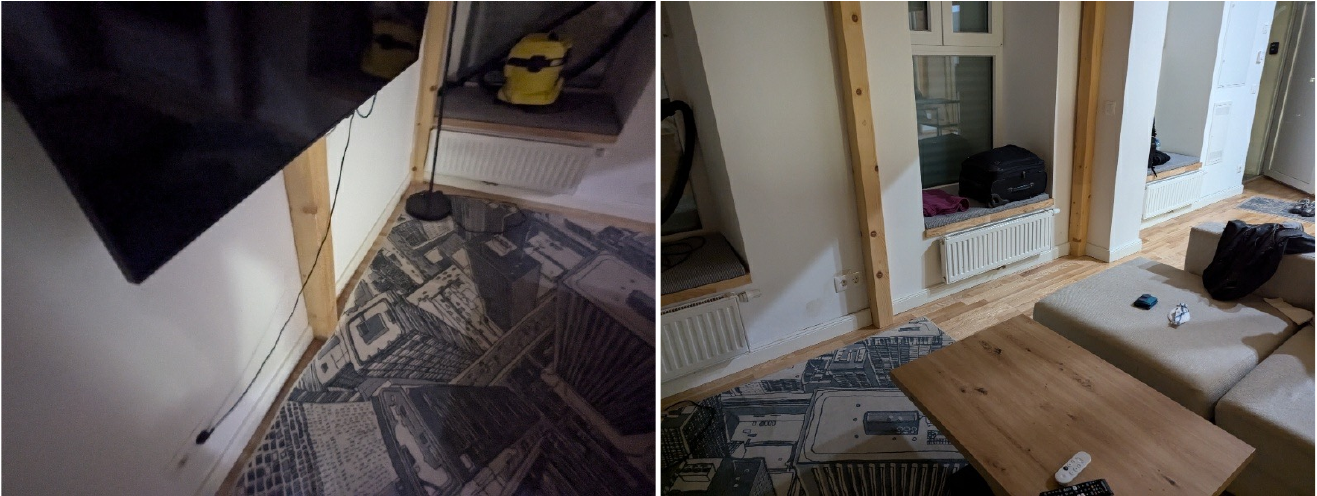}
\\
\textbf{Prompt} & What object is located immediately to the right of point [710 991] in the second image, just outside of the frame ?
\\\cmidrule{1-2}
\textbf{Answer} & Wooden Table
\\
\bottomrule
\end{tabular}

\captionof{table}{Examples of \textit{L3.Causal Reasoning}.}
\label{tab:example_3dod_data}
\end{minipage}
\end{table*}

\begin{table*}[t]
\begin{minipage}{1.0\linewidth}
\centering
\tablestyle{4.0pt}{1.0}

\begin{tabular}{p{0.11\linewidth} p{0.85\linewidth}}
\toprule
 \multicolumn{2}{l}{\bf Sequential Planning}  \\
\midrule
&  \includegraphics[width=0.7\textwidth]{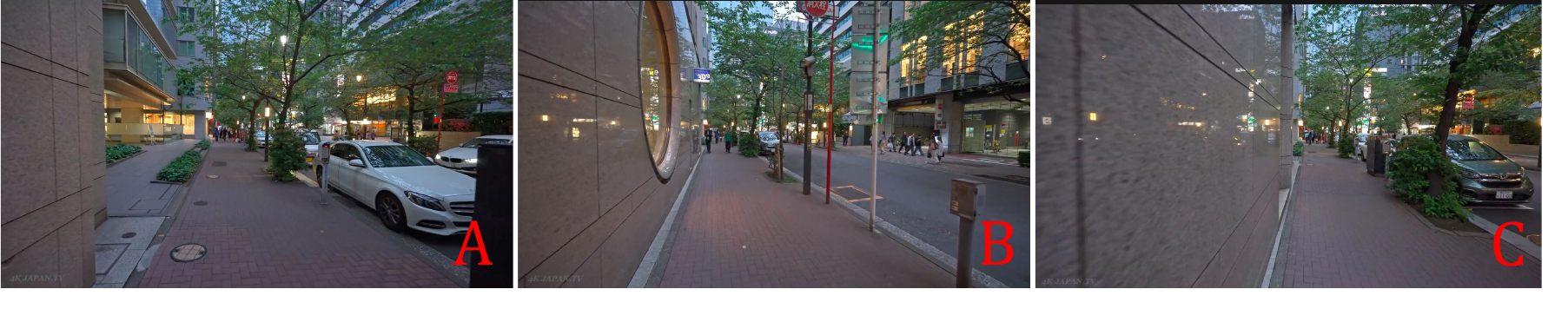}
\\
\textbf{Prompt} & The camera is moving forward. Please arrange these three images in chronological order?
\\\cmidrule{1-2}
\textbf{Answer} & B-C-A
\\\midrule
&  \includegraphics[width=0.27\textwidth]{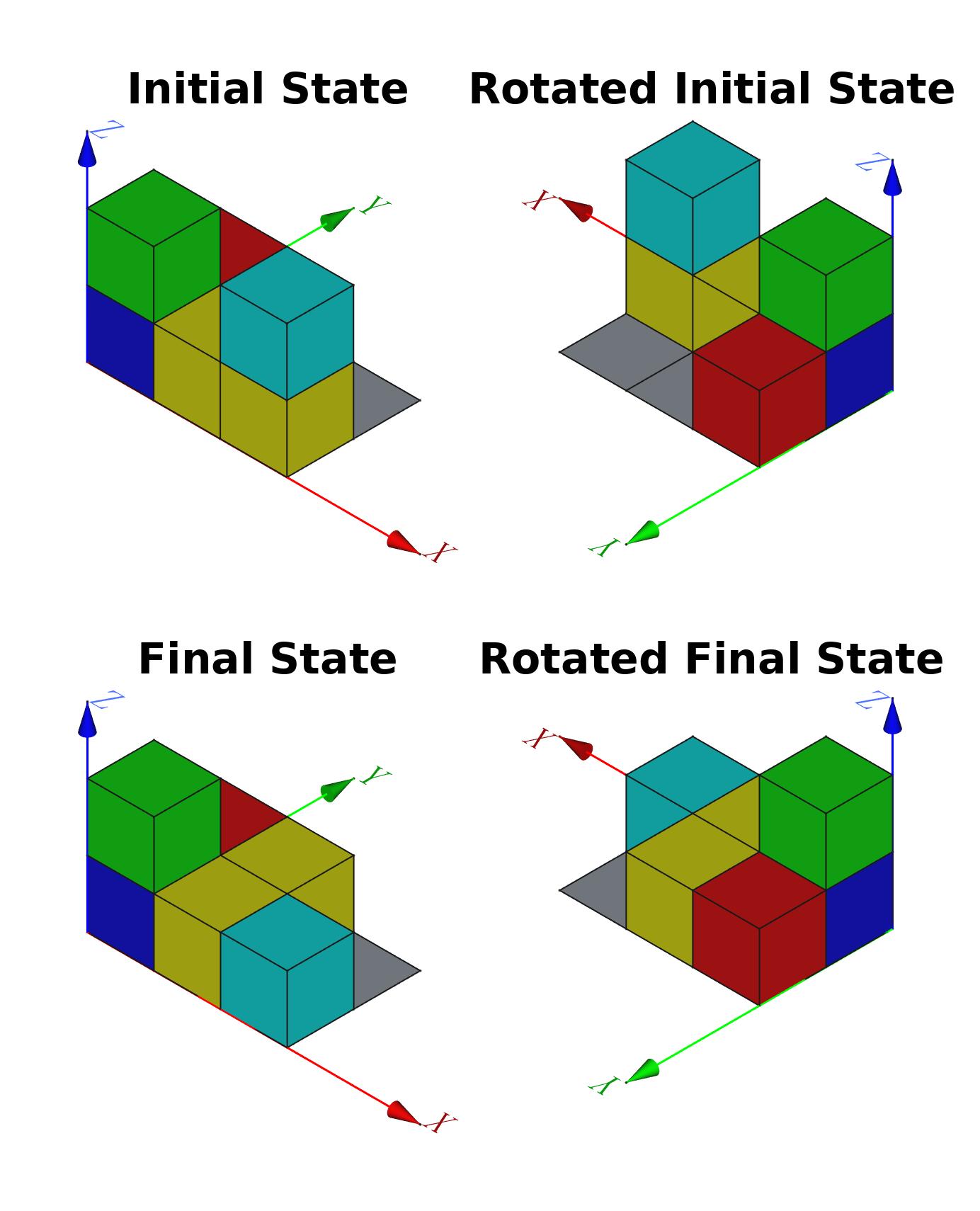}
\\
\textbf{Prompt} & The top row of images shows different views of the initial state of a cube stack, while the bottom row shows different views of the final state after transformation. During the transformation process, blocks can move one unit in any direction (forward, backward, left, right, up, down). If the target position is empty, the block can move there directly; if the target position already has a block, they swap places. Blocks cannot float in the air. If a block is moved away from a position, any block above it will fall down until reaching a supporting surface. The xyz axes are shown in the diagram, and each block's position can be precisely identified using coordinates (x1,y1,z1). Which of the following transformation sequences can change the cube stack from the initial state to the final state shown in the diagram? Please answer from options A, B, C, or D.
\begin{verbatim}
A. (1, 0, 0) y+ -- (0, 0, 1) z-
B. (1, 0, 0) x+ -- (1, 0, 0) y+
C. (2, 0, 0) x- -- (1, 0, 0) y+ -- (2, 0, 0) x-
D. (0, 0, 0) x+ -- (0, 1, 0) y- -- (0, 0, 1) y+
\end{verbatim}
\\\cmidrule{1-2}
\textbf{Answer} & 
\begin{verbatim}
	C
\end{verbatim}
\\
\bottomrule
\end{tabular}

\captionof{table}{Examples of \textit{L3.Sequential Planning}.}
\label{tab:example_3dod_data}
\end{minipage}
\end{table*}

\begin{table*}[t]
\centering
\tablestyle{4.0pt}{1.0}

\begin{tabular}{p{0.11\linewidth} p{0.85\linewidth}}
\toprule
\multicolumn{2}{l}{\bf L4 Agentic Competence} \\
\midrule

& \includegraphics[width=0.8\textwidth]{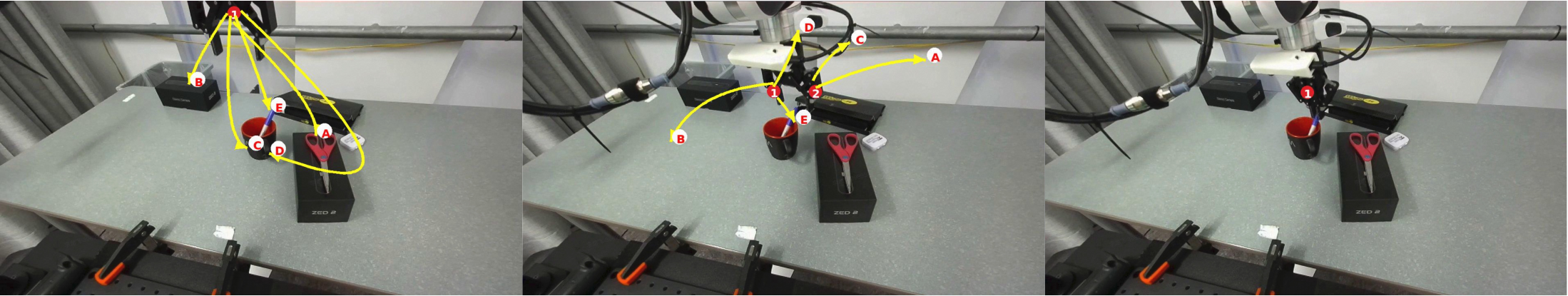}
\\
\textbf{Prompt} & You are an intelligent agent observing a video sequence that depicts a task being performed which is "please get the blue pan out of the bottle".
For each image in the sequence that provides candidate action options, select the single most appropriate action to perform in the current state.
In the final frame, do not select any action. Instead, determine whether the overall task shown in the sequence has been successfully completed (1) or not completed (0).
Please output the selected action for each intermediate frame and the completion flag for the final frame.
\begin{verbatim}
A. EB1
B. CC1
C. EA0
D. EE1
E. AD0
F. EE0
G. CD0
H. EC0
\end{verbatim}
\\ \cmidrule{1-2}
\textbf{Answer} & F
\\ \midrule

& \includegraphics[width=0.7\textwidth]{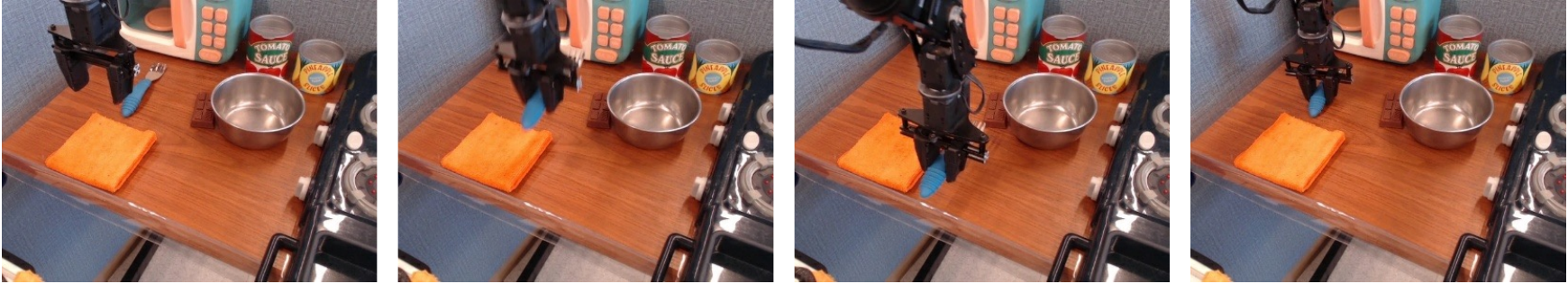}
\\
\textbf{Prompt} & Which image shows the robot making the most progress towards the task placing the fork to the right side of the orange kitchen wipe?
\begin{verbatim}
A. First.
B. Second.
C. Third.
D. Fourth
\end{verbatim}
\\ \cmidrule{1-2}
\textbf{Answer} & 
\begin{verbatim}
	C
\end{verbatim}
\\
\bottomrule
\end{tabular}
\caption{Examples of \textit{L4 Agentic Competence}.}
\label{tab:example_3dod_data_part1}
\end{table*}

\begin{table*}[t]
\centering
\tablestyle{4.0pt}{1.0}

\ContinuedFloat

\begin{tabular}{p{0.11\linewidth} p{0.85\linewidth}}
\toprule
\multicolumn{2}{l}{\bf L4 Agentic Competence (Continued)} \\
\midrule

& \includegraphics[width=0.7\textwidth]{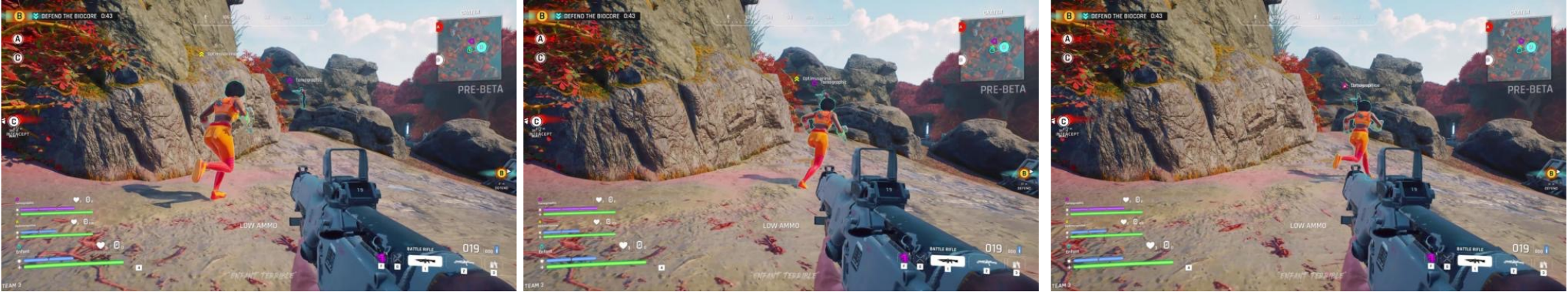}
\\
\textbf{Prompt} & 
\textbf{Task: Visual Navigation Action Sequence Generation}

You are an expert visual navigation agent. Your task is to generate a sequence of actions to navigate a robot from a starting visual state (Image 2) to a target visual state (Image 3) based on the provided visual information.

\vspace{0.5em}
\textbf{Context and Example}

We provide three sequential images: Image 1, Image 2, and Image 3.
To help you understand the task, we are providing the complete action sequence that navigates from \textbf{Image 1} to \textbf{Image 2} as an example.

\textit{Example Action Sequence from Image 1 to Image 2:}
\begin{verbatim}
{
  "actions": ["Dolly In", "Truck Left", "Pedestal Up", 
              "Pan Left", "Roll CW"], 
  "step_nums": [0, 3, 4, 5, 0]
}
\end{verbatim}

\textbf{Your Core Task}

Now, carefully observe \textbf{Image 2} (the starting state) and \textbf{Image 3} (the target state). Your mission is to generate the action sequence required to navigate from Image 2 to Image 3.

\vspace{0.5em}
\textbf{1. Action Space}

You must choose from the following 12 elementary actions. In your output, you \textbf{must use the `symbol' specified} to represent each action.

\vspace{0.5em}
{\footnotesize
\begin{tabular}{@{}llcl@{}}
\toprule
\textbf{Category} & \textbf{Action} & \textbf{Sym} & \textbf{Description} \\
\midrule
Trans. & Dolly In & W & Move forward \\
 & Dolly Out & S & Move backward \\
 & Truck Left & A & Move left \\
 & Truck Right & D & Move right \\
 & Pedestal Up & space & Move up \\
 & Pedestal Down & shift & Move down \\
\midrule
Rot. & Pan Left & $\leftarrow$ & Neg. rot around +Y \\
 & Pan Right & $\rightarrow$ & Pos. rot around +Y \\
 & Tilt Up & $\uparrow$ & Pos. rot around +X \\
 & Tilt Down & $\downarrow$ & Neg. rot around +X \\
 & Roll CW & $\circlearrowright$ & Neg. rot around +Z \\
 & Roll CCW & $\circlearrowleft$ & Pos. rot around +Z \\
\midrule
Special & Stay & STOP & No movement \\
\bottomrule
\end{tabular}
}

\vspace{0.5em}
\textbf{2. Step Size Parameters}
The magnitude of each action is determined by \texttt{step\_num} combined with a unit step size.
\begin{itemize}[leftmargin=*, nosep]
    \item \textbf{Translation}: 0.0626 meters per step.
    \item \textbf{Rotation}: 0.0725 radians per step.
\end{itemize}
E.g., \texttt{action\_symbol: "W"} and \texttt{step\_num: 10} means moving forward by $10 \times 0.0626$ meters.

\vspace{0.5em}
\textbf{3. Required Output Format}
Your final output \textbf{MUST} be a JSON object containing two keys: \texttt{"actions"} and \texttt{"step\_nums"}. The lengths of both arrays must be identical.
\\ \cmidrule{1-2}
\textbf{Answer} & 
\begin{verbatim}
{
  "actions": [
    "Truck Left",
    "Pedestal Up",
    "Pan Left"
  ],
  "step_nums": [
    1,
    4,
    1
  ]
}
\end{verbatim}
\\
\bottomrule
\end{tabular}
\caption{Examples of \textit{L4 Agentic Competence} (Continued).}
\label{tab:example_3dod_data_part2}
\end{table*}

\end{document}